\pdfoutput=1

\documentclass[11pt]{article}

\usepackage[final]{acl}

\usepackage{times}
\usepackage{latexsym}
\usepackage[T1]{fontenc}
\usepackage[utf8]{inputenc}

\usepackage{microtype}
\usepackage{inconsolata}

\usepackage{graphicx}
\usepackage{tabularx}
\usepackage{url}
\usepackage{wrapfig}
\usepackage{amsmath}
\usepackage{amssymb}
\usepackage{bbm}
\usepackage{mdframed}
\usepackage{multicol}
\usepackage{soul}
\usepackage{subcaption}
\usepackage{booktabs}
\usepackage{multirow}
\usepackage[nice]{nicefrac}
\usepackage{placeins}
\usepackage{enumitem}
\usepackage{tcolorbox}
\usepackage{pdflscape}
\usepackage{minitoc}

\usepackage{rotating}
\usepackage{listings}
\usepackage{mathtools} 
\usepackage{adjustbox}
\usepackage{comment}
\usepackage{pifont}
\usepackage[boxed]{algorithm2e}
\usepackage[accsupp]{axessibility}
\usepackage[fixed]{fontawesome5}
\usepackage{authblk}
\usepackage{natbib}

\definecolor{kleinblue}{RGB}{0, 47, 167} 
\definecolor{kleinblue2}{RGB}{20, 20, 125} 
\definecolor{kleinred}{HTML}{bc1919}
\definecolor{cvprblue}{rgb}{0.21,0.49,0.74}
\definecolor{softgreen}{rgb}{0.13, 0.55, 0.13}
\definecolor{softred}{rgb}{0.8, 0.13, 0.13}

\usepackage{cleveref}[2012/02/15]

\newcommand{\stoptocwriting}{%
  \addtocontents{toc}{\protect\setcounter{tocdepth}{-5}}}

\title{\textsc{ONEBench} to Test Them All:\\ Sample-Level Benchmarking Over Open-Ended Capabilities}

\author{
  Adhiraj Ghosh\textsuperscript{1,2}\thanks{Equal contribution, random order, $^\circ$ core contributors}\quad
  Sebastian Dziadzio\textsuperscript{1,2*}\quad
  Ameya Prabhu\textsuperscript{1,2$^\circ$}\quad
  Vishaal Udandarao\textsuperscript{1,2,3$^\circ$}
  \\ \vspace{0.5ex}
  Samuel Albanie \quad
  Matthias Bethge\textsuperscript{1,2}
  \\\vspace{1ex}
  \textsuperscript{1}Tübingen AI Centre, University of Tübingen \quad
  \textsuperscript{2}Open-$\Psi$ Collective \quad
  \textsuperscript{3}University of Cambridge\\
  \texttt{\href{mailto:adhiraj.ghosh@bethgelab.org}{adhiraj.ghosh@bethgelab.org}}
\\
\\
  \noindent\faGlobe\ \href{https://bethgelab.github.io/onebench/}{\texttt{Project Page}} \quad
  \faGithub\ \href{https://github.com/bethgelab/onebench/}{\texttt{Code}} 
}

\begin{document}
\maketitle

\begin{abstract}
Traditional fixed test datasets fall short in evaluating the open-ended capabilities of foundation models. To address this, we propose ONEBench (\textbf{O}pe\textbf{N}-\textbf{E}nded \textbf{Bench}marking), a new paradigm that consolidates individual evaluation datasets into a unified, ever-expanding sample pool. ONEBench enables custom benchmarks for specific capabilities while reusing and aggregating samples, mitigating overfitting and dataset bias for broader capability assessment. It reframes model evaluation as selecting and aggregating sample-level tests. Transitioning from task-specific benchmarks to ONEBench introduces two challenges: \emph{heterogeneity} (aggregating diverse metrics) and \emph{incompleteness} (comparing models tested on different data subsets).
To address these, we propose an aggregation algorithm that ensures identifiability---asymptotically recovering ground-truth scores---and rapid convergence, enabling accurate model comparisons with relatively little data. On homogenous datasets, our algorithm produces rankings that highly correlate with average scores. Moreover, it remains robust to over $95$\% missing measurements, reducing evaluation costs by up to 20 times. We introduce ONEBench-LLM for language models and ONEBench-LMM for vision-language models, enabling targeted model testing across diverse capabilities. 
\end{abstract}    
\section{Introduction}
\label{Introduction}
Deep learning has arrived in the post-dataset era\footnote{\href{https://www.youtube.com/watch?v=hJGJF32idMU}{From a talk by Alexei Efros at ICML 2020}}. As foundation models rapidly expand their zero-shot capabilities, the focus of evaluation has moved beyond singular, dataset-specific performance measurements that rely on dividing fixed collections of data into train and test sets. Instead, foundation models are employed as general knowledge and reasoning engines across a wide range of domains. This creates a pressing need to characterize their open-ended capabilities using diverse metrics in zero-shot settings~\citep{ge2024openagi}. However, static benchmarks, which test generalization on fixed test splits, cannot probe the ever-evolving capabilities of foundation models effectively. This raises an important question: \textit{How can benchmarking adapt to measure an open-ended set of capabilities?}

\begin{figure*}[t]
\centering
   \includegraphics[width=0.9\linewidth]{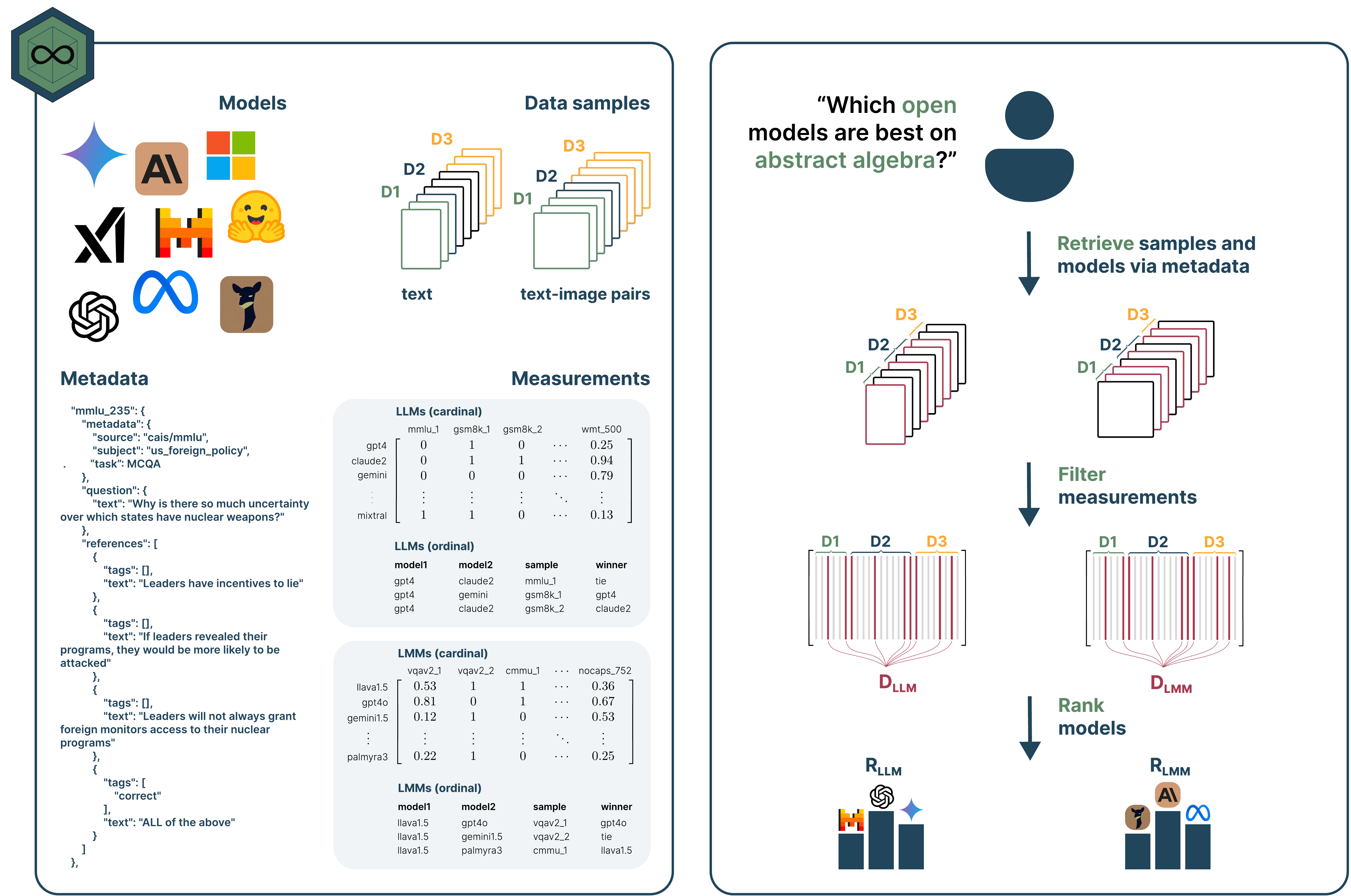}
   \caption{\textbf{The ONEBench Framework.} \textit{Left}: ONEBench comprises a set of models, a pool of data samples spanning multiple test sets, metadata describing models and data samples, and a collection of sample-level measurements. \textit{Right}: the user formulates a query to capture the desired model capability, using a mix of structured metadata filters and semantic search. Selected models are then ranked on a subset of data samples that meet the specified criteria.}
   \label{fig:benchmark_overview}
\end{figure*}

We propose a solution based on dynamic, sample-level evaluation, which we call 
ONEBench (\textbf{O}pe\textbf{N}-\textbf{E}nded \textbf{Bench}marking). In this approach, test sets for particular capabilities are generated ad-hoc from a large pool of individual annotated data samples. These sample-level evaluations act as atomic units of measurement that can be flexibly aggregated into an exponential number of configurations. Thanks to this flexibility, the sample pool and corresponding annotation metrics can be continuously updated to incorporate new evaluations. Additionally, this approach can reduce \textit{dataset bias}---systematic quirks in the data arising from its collection process~\citep{liu2024decade}.
Finally, by combining samples across test sets, ONEBench  captures real-world diversity~\citep{ni2024mixeval}.

The most important feature of ONEBench is its potential to democratize evaluation. Unlike traditional benchmarks, typically created by individual groups based on their own criteria for data collection and evaluation procedures~\cite{ bansal2024peekingbehindclosed}, ONEBench integrates test sets from multiple sources reflecting a wide range of perspectives, use cases, and objectives. This flexibility allows different interest groups to collaboratively define their own evaluations by selecting the most appropriate combination of tests that best suit their specific requirements. Moreover, the design of ONEBench challenges the dominant approach of chasing single benchmark scores, which fail to account for the difficulty of individual data instances~\citep{ethayarajh2022understanding}, in favor of a plurality of rankings and a dynamic, granular, multi-faceted evaluation.

\textcolor{kleinblue2}{\textbf{Challenges.}} Building ONEBench requires addressing two key challenges: \textit{heterogeneity} and \textit{incompleteness}. \textit{Heterogeneity} arises because model evaluations span diverse metric types, such as binary (correct/incorrect), numeric (BLEU scores), and ordinal (preference rankings), making aggregation difficult. \textit{Incompleteness} occurs when models are tested on non-overlapping subsets of data, preventing fair and direct comparisons. Traditional benchmarks sidestep these issues by using a multi-task setup, where all models are evaluated on the same samples using a single metric.

\textcolor{kleinblue2}{\textbf{Solution and Theoretical Guarantees.}} We address these challenges using social choice theory, treating data samples as voters expressing preferences over models. By converting all measurements into ordinal rankings, we leverage established principles to robustly aggregate heterogeneous and incomplete data. Our approach assumes a random utility model based on the Plackett-Luce framework~\citep{plackett1975analysis,luce1959individual}, which provides guarantees for accurately recovering ground-truth utility scores. This approach ensures that our model rankings are both theoretically sound and practical, with rapid convergence guarantees enabling accurate rankings from limited data.

\textcolor{kleinblue2}{\textbf{Why rankings?}} While converting cardinal measurements to ordinal ones results in information loss, two key considerations justify this choice in benchmarking: \textit{external validity} and \textit{benchmark longevity}. \textit{External validity}~\citep{liao2021we} refers to how well evaluation results generalize across different settings. \citet{recht2019imagenet} found that although accuracy shifts across test sets, model rankings remain stable. \citet{salaudeen2024imagenot} extended this by comparing ImageNet models to those trained and tested on a new dataset, arguing that ordinal comparisons are more robust, e.g., injecting random labels into two otherwise identical benchmarks causes accuracy to vary while preserving relative order. Given the noise in LLM and LMM evaluation (from prompt design, label choices, in-context examples, etc.), rankings offer a more reliable signal than absolute values. \textit{Benchmark longevity} is another growing challenge, as internet-scale datasets lead to faster saturation. Contamination and adaptive overfitting threaten evaluation integrity. While ONEBench mitigates contamination by filtering compromised samples, its use of rankings helps guard against overfitting---an idea inspired by the Ladder algorithm~\citep{hardt2022patterns}.

\textcolor{kleinblue2}{\textbf{Empirical Validation.}} We create ONEBench for two domains: ONEBench-LLM for language models and ONEBench-LMM for vision-language models. These benchmarks unify evaluations by aggregating data from diverse sources, including preference data (arenas) and heterogeneous multi-task leaderboards.
Our empirical results demonstrate that the Plackett-Luce model effectively aggregates real-world benchmarks, showing a high correlation with ground-truth score-based rankings over homogeneous datasets. Notably, this strong correlation persists even when up to $95$\% of the data is missing, enabling a 20 times reduction in evaluation costs with minimal impact on performance. Finally, we compare Plackett-Luce rankings to widely adopted methods such as ELO~\cite{elo1967proposed} and Bradley-Terry~\citep{bradley1952rank}, demonstrating superior accuracy and robustness to missing data. We additionally compare our Plackett-Luce rank aggregation approach to popular computational social choice theory methods like Borda Count and the Dowdall System (\cref{subsec:social_choice}).

\textcolor{kleinblue2}{\textbf{Personalized Aggregation.}} Imagine you are a biochemist seeking an LLM to assist with designing experiments related to antibodies. With ONEBench, you can input a query, such as ``\texttt{immunology}'' or ``\texttt{antibodies}'' to generate a dynamically constructed benchmark that ranks models based on their performance in \textit{this specific domain}. While the optimal selection of personalized capability sets remains an open research challenge, we present a proof of concept by distinguishing between \emph{tasks} (e.g., reading comprehension) and \emph{concepts} (e.g., Clostridium bacteria). By combining structured filters and flexible semantic search, users can define their capability of interest along these dimensions and conduct targeted evaluations, resulting in personalized rankings.

To summarize, ONEBench is a democratized, open-source collection of diverse evaluation samples enriched with detailed metadata. Its robust aggregation method ranks models across heterogeneous metrics and incomplete evaluation data. Users can perform semantic searches and apply structured query filters to dynamically generate benchmarks tailored to their needs. They can also contribute new evaluation samples and model measurements, which are instantly aggregated to refine rankings. This framework enables lifelong aggregation of arbitrary test sets with unprecedented flexibility and precision.

\section{ONEBench: Formulation}


\subsection{Components}
\label{subsec:formalisation}
The goal of ONEBench is to evaluate a set of models $\{m_k\}_{k=1}^{M}$ using a continuosly expanding pool of test data samples $\mathcal{D}$ drawn from multiple benchmarks $\{\mathcal{B}_k\}_{k=1}^{B}$. Each data sample may include metadata specifying the capabilities it is testing. To handle the diversity of data from different benchmarks, we generate sample-level rankings (${\mathcal{S}}$) for all samples in the test pool. \Cref{fig:benchmark_overview} provides a schematic overview of ONEBench, with each component described below.

\noindent\textcolor{kleinred}{\textbf{(i) Data Pool.}} The data pool $\mathcal{D} = \{(x_k, y_k)\}_{k=1}^D$ consists of data samples $x_k$ with reference answers $y_k$. An example of a data sample is the question \textit{``What was the dominant strain of flu in 2010? Select among four choices."} with a reference answer \textit{``H1N1/09"}. Each instance can also include metadata specifying tested capabilities, for example as a list of keywords like \textit{temporal QA, pandemics, history, biology, virology, multiple-choice QA}.
 
\noindent\textcolor{kleinred}{\textbf{(ii)  Models.}} The set of models is defined as $\mathcal{M} = \{m_{base}\} \cap \{m_k\}_{k=1}^{M}$, where $m_{base}$ serves as a baseline for evaluating the capabilities of the other models. A common choice for $m_{base}$ is a random model.  Since the original benchmarks evaluate different sets of models, each benchmark ${\mathcal{B}_k}$ considers a subset of models ${\mathcal{M}_{\mathcal{B}_k} } \subseteq \mathcal{M}$.
 
\noindent\textcolor{kleinred}{\textbf{(iii) Sample-level Rankings.}} For each data sample $(x_j, y_j) \in \mathcal{D}$, we construct a sample-level ranking ${s_j \in \mathcal{S}}$ over model subset ${\mathcal{M}_j \subseteq \mathcal{M}_{\mathcal{B}_{k}} }$, where $k$ denotes the index of the benchmark from which the sample $(x_j, y_j)$ was collected (more in \cref{appx:rankings}). Crucially, these rankings depend only on the evaluation metrics used by each benchmark, abstracting away the specifics of those metrics. This abstraction is central to our approach, as it enables aggregation across heterogeneous evaluation metrics.

\noindent\textcolor{kleinred}{\textbf{(iv) Capabilities.}} To enable selective retrieval of relevant sample-level rankings in $\mathcal{B}$ based on user queries, each ranking can be associated with a \textit{capability}. We distinguish between \textit{tasks} (e.g., question answering, captioning) and \textit{concepts} (e.g., makeup, geometry). Since capabilities are inherently open-ended, we only tag data samples with task information, while concept-based retrieval is performed dynamically at test time using semantic search.

\textcolor{kleinblue2}{\textbf{Lifelong Expansion.}}
Data pool $\mathcal{D}$ and model set $\mathcal{M}$ are stored as tables, while sample-level model evaluations are represented as a relational database linking these tables. Expanding ONEBench over time requires augmenting $\mathcal{D}$, $\mathcal{M}$, and $\mathcal{S}$ through the following operations: $\texttt{insert}_{\mathcal{D}}, \texttt{insert}_{\mathcal{M}}, \texttt{insert}_{\mathcal{S}}$. The first two operations simply add new samples and models, while $\texttt{insert}_{\mathcal{S}}$ registers a new sample-level ranking.

\subsection{Capability Querying}
\label{cap-probing-section}
To evaluate a given capability, ONEBench takes a dynamic approach. First, we retrieve samples that match the query. Then, we aggregate the sample-level rankings to produce the overall ranking.

\noindent\textcolor{kleinred}{\textbf{Retrieve ($\texttt{retrieve}_\mathcal{D}$).}} Here, the system selects relevant data instances based on a user's query. The query language is flexible and allows retrieving data instances that semantically relate to a specific topic or match certain criteria. Retrieval combines k-nearest neighbors (kNN) search over dense embeddings, with the query as input, and structured queries that leverage the unified data schema.

\noindent\textcolor{kleinred}{\textbf{Aggregate ($\texttt{Aggregate}_{\mathcal{S},\mathcal{D}}$).}} Measurements over the retrieved subset are combined using the random utility modelling approach \citep{xia2019learning}, defining a joint probability distribution over all measurements (sample rankings $s_j$ and model scores $\gamma_j)$, given model permutations $\sigma_j$ and binary sequence of pairwise performance relations $\pi_j$  assuming statistical independence:
 \vspace{-0.3cm}
\begin{multline}
p(s_1, \dots, s_{n_\infty}|\gamma_1, \dots, \gamma_M) = \\
\prod_{j=1}^{n_\infty} p(s_j =[.]_{(\sigma_j, \pi_j)}|\gamma_1, \dots, \gamma_M).
\nonumber
\end{multline}

\noindent The Placket-Luce framework assumes the following probability model:
\vspace{-0.3cm}
\begin{multline}
p\left(s_j=[.]_{(\sigma_j, \pi_j)}
\right) = \\
\frac{\gamma_{\sigma_j(1)}}{\underbrace{\sum_{k=1}^{m_j} \gamma_{\sigma_j(k)}}_{f_{\sigma_j(1)}}} \times 
\cdots \times \frac{\gamma_{\sigma_j(m_j-1)}}{\underbrace{\gamma_{\sigma_j(m_j-1)} + \gamma_{\sigma_j(m_j)}}_{f_{\sigma_j(m_j)}}}   
\nonumber,
\end{multline}
\vspace{-0.3cm}

defining one parameter $\gamma_k$ for each model $m_k$ that determines its performance relative to all other models. 
To aggregate model performances over sample rankings, we estimate parameters

\vspace{-0.5cm}
\begin{equation*}
    \hat{\mathbf{\gamma}} = \operatorname*{argmax}_{\mathbf{\gamma} \in \mathbb{R}^m} 
    \log p(\mathbf{s} \mid \mathbf{\gamma})
\end{equation*}
\vspace{-0.3cm}

\noindent with maximum likelihood estimation (MLE). The global ranking follows the permutation $\sigma_\infty$ where
$\hat \gamma_{\sigma_\infty(1)} > \dots > \hat \gamma_{\sigma_\infty(m)}$. The ML condition uniquely determines all performance parameters $\{\hat{\gamma}_k\}_{k=1}^M$, as the likelihood function is strictly concave. The parameters of the Plackett-Luce model are identifiable up to an arbitrary additive constant. Consistency and asymptotic normality can also be shown under certain assumptions about the comparison graph \citep{han2023unified}. We refer to the estimated latent variables $\{\hat{\gamma}_k\}_{k=1}^M$ as {\it model scores}. A model with a higher score likely performs better on a randomly picked sample-level task than one with a lower score. To fix the additive constant, we set the baseline model score  $\hat \gamma_{baseline}$ to zero.

\section{ONEBench: Aggregation}
\label{subsection:ranking}
\vspace{-0.1cm}
We view aggregating sparse ordinal preferences over models through a computational social choice lens, where samples are voters, models are candidates, and the aggregation algorithm is the voting mechanism~\citep{brandt2016introduction}. 
We aggregate ordinal comparisons with partial data to produce a global ranking and analyze its properties.

\subsection{Theoretical Foundations}
We begin by postulating a ground-truth statistical model generating the data, which is converted into ordinal comparisons ($\mathcal{S}$). This is in contrast to~\citet{zhang2024inherent}, who view aggregation as classical voting and analyse tradeoffs in aggregating voter preferences rather than uncover an underlying ranking. Specifically, we use a random-utility model \citep{thurstone1927three}, where model $m_i$ is associated with utility distribution $\mathcal{U}_{m_i}$. Preferences between models $m_i$ and $m_j$ are based on comparing sampled utilities, i.e., $m_i \prec m_j:= u(m_i)<u(m_j)$, where $u_{m} \sim \mathcal{U}_{m}$. Since computing maximum likelihood estimates over general random-utility models is computationally hard \citep{xia2019learning}, we focus on the Plackett–Luce model \citep{plackett1975analysis,luce1977choice}, the only known exception that allows for tractable MLE.

\textcolor{kleinblue2}{\textbf{Property 1: Identifiability.}}
We first ask: \textit{Are the utility distributions for all models recoverable?} The Plackett-Luce model allows identifying the utility distribution (up to an arbitrary additive constant) if all models are compared via a directed path \citep{xia2019learning}. Using reference model $m_{\text{base}}$ removes additive ambiguity. Consistency and asymptotic normality hold under specific assumptions about the comparison graph \citep{han2023unified}.

\textcolor{kleinblue2}{\textbf{Property 2: Sample-Efficient Convergence from Sparse Data.}} Given that identifiability is asymptotic, we ask: \textit{How sample-efficient is the algorithm for recovering the utility distribution?} With partial rankings of size $k$, the MLE is  surprisingly sample efficient while being minmax-optimal \citep{maystre2015fast}. 
Sampling $k$ model comparisons from the model set $|\mathcal{M}|$ uniformly at random induces an expander graph with high probability, giving guarantees for sample-efficient recovery,
with $\nicefrac{\Omega(|\mathcal{M}|)}{k}$ samples being necessary, and $\nicefrac{\Omega(|\mathcal{M}|\log|\mathcal{M}|)}{k}$ samples being sufficient. Efficient algorithms like 
\citet{maystre2015fast} achieve these bounds. Rank-breaking techniques, used in our evaluation, offer near-optimal solutions \citep{soufiani2014computing}.

\textcolor{kleinblue2}{\textbf{Property 3: Social Properties.}} The Plackett-Luce model offers computational efficiency and recoverability of the underlying ranking. However, designing democratic decision-making systems also requires fair aggregation. Ensuring fairness involves trade-offs~\citep{zhang2024inherent}, as different notions of fairness often conflict. Moreover, depending on the intended application areas, differing or even opposing preferences may be valid \citep{arrow1950difficulty}.
Plackett-Luce offers ``procedural fairness" (\citealp{sep-social-choice}), satisfying:

\textcolor{kleinblue2}{\textbf{(i) Anonymity.}} All voters (samples) are treated equally, ensuring the system does not over-rely on any single vote. Rankings remain unchanged if the input sample set is permuted. 

\textcolor{kleinblue2}{\textbf{(ii) Neutrality.}} The ranking is invariant to model identities, ensuring fairness among alternatives. This means permuting the models similarly permutes the resulting ranking.

\textcolor{kleinblue2}{\textbf{(iii) Independence from Irrelevant Alternatives.}} The relative ranking of two models is unaffected by other alternatives in a given sample, as guaranteed by~\citet{luce1959individual}. This provides grounding for incomplete model evaluations.  

In Section~\ref{sec:limitations}, we discuss separability and pairwise majority consistency---two properties violated by the Plackett-Luce model.

\subsection{Translating Theory to Practice}
\begin{figure*}[t]
\centering
    \includegraphics[width=\textwidth]{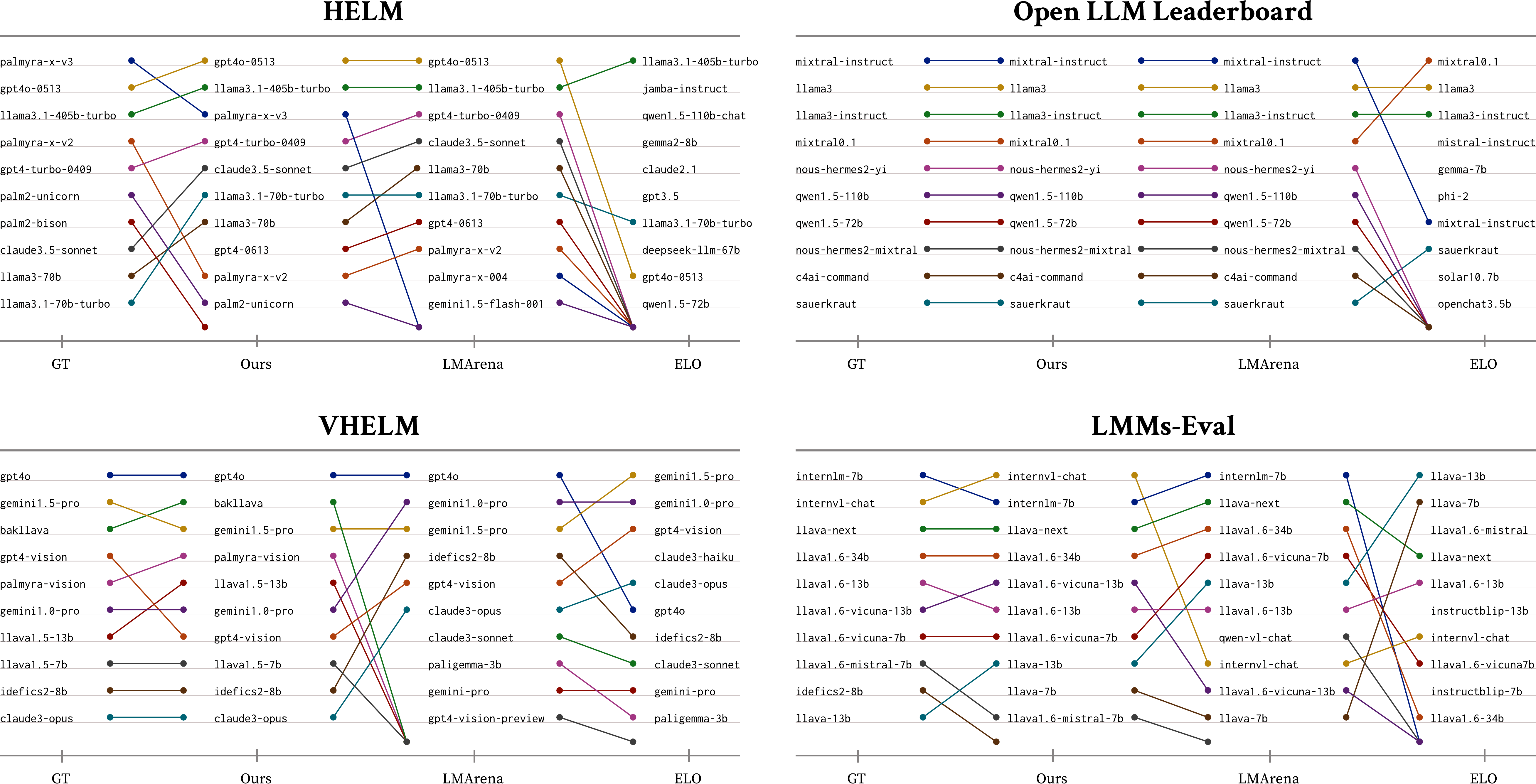}
    \caption{\textbf{Top-10 model ranking changes across different aggregation methods.} Plackett-Luce (\texttt{Ours}) shows the most similarity to the Ground Truth model rankings (\texttt{GT}). However, there is a progressive degradation in ranking accuracy for LMArena  (\texttt{LMArena}) and Elo (\texttt{ELO}).}
    \label{fig:rank_change}
\end{figure*}

Here, we show that: (i) the Plackett-Luce model works well on real-world data, (ii) our aggregation method is sample-efficient, and (iii) it handles high levels of incompleteness. 
Below, we describe our setup and address these points.

\subsubsection{Setup}
\label{subsec:setup}
\noindent \textcolor{kleinred}{\textbf{Benchmarks.}} We conduct experiments using four popular benchmarks with established model rankings based on benchmark-specific average scores: HELM~\citep{liang2022holistic} and Open LLM Leaderboard~\citep{open-llm-leaderboard} for LLMs, and VHELM~\citep{VHELM2024} and LMMs-Eval~\citep{zhang2024lmms} for LMMs.  We 
define our data pool as the sum of all samples in the constituent datasets. To test the faithfulness of our aggregation strategy we compare the resulting rankings to the original leaderboards. These leaderboards evaluate models across varied tasks with different metrics, serving as good indicators of real-world performance.

\noindent \textcolor{kleinred}{\textbf{Ground Truth.}} The current system of benchmarking involves evaluating models on individual test sets and measuring the mean score per model. This holds even for benchmarks that combine test sets.
We consider these scores as the ground truth measurement and generate a ground truth model ranking from these scores. Since we aggregate multiple measurement metrics, we implement a min-max normalization of numeric measurements to bring all benchmark samples to the same 0-1 score range. Our final ground truth refers to the model rankings derived from the mean score across all benchmarks. While this procedure is suboptimal, and indeed we propose to replace it with our method, there is no other ground truth available for real data. We show results on synthetic data in~\cref{appx:synthetic}.

\noindent\textcolor{kleinred}{\textbf{Methods.}} We evaluate three ranking methods:

\textcolor{kleinblue2}{\textbf{(i) Elo Score}} \citep{elo1967proposed}: A competitive game rating system adapted to rank models through pairwise comparisons, adjusting scores based on wins or losses to reflect win-rate reliability.

\textcolor{kleinblue2}{\textbf{(ii) LMArena Ranking}}: A ranking method based on the Bradley-Terry model~\citep{bradley1952rank}, using a Maximum Likelihood Estimation (MLE) based on pairwise comparisons with an underlying ELO model for rank aggregation.

\textcolor{kleinblue2}{\textbf{(iii) Ours}}: We leverage the Plackett-Luce model~\citep{maystre2015fast} to aggregate pairwise comparisons using partial rank breaking, speeding up rank estimation.

\noindent \textcolor{kleinred}{\textbf{Metrics.}} We compare the rankings generated by each method to the ground-truth from the leaderboards using Kendall’s $\tau$, a standard correlation metric for rankings. Each method is tested three times and we report the mean and variance. We also check that the top-$k$ models are reliably recovered.

\begin{table}[htbp]
\centering
\resizebox{\linewidth}{!}{
\begin{tabular}{lccc}
\toprule
\textbf{Dataset} & \textbf{Elo} & \textbf{LMArena} & \textbf{Ours} \\ \midrule
HELM & 0.35 ± 0.13 & 0.85 ± 0.00 & \textbf{0.88 ± 0.00} \\
Leaderboard & 0.21 ± 0.07 & 0.97 ± 0.00 & \textbf{0.99 ± 0.00} \\
VHELM & 0.63 ± 0.02 & 0.69 ± 0.00 & \textbf{0.80 ± 0.00} \\
LMMs-Eval & 0.33 ± 0.11 & 0.42 ± 0.00 & \textbf{0.64 ± 0.00} \\ 
\bottomrule
\end{tabular}
}
\caption{\textbf{Comparison to other aggregation algorithms.} Kendall's $\boldsymbol{\tau}$ correlations to ground-truth ranking across individual benchmarks.}
\label{tab:gt_results}
\end{table}

\subsubsection{Plackett-Luce on Real-World Data}

\textcolor{kleinblue2}{\textbf{Q1. Is it suitable?}}  We evaluate the Plackett-Luce model on large-scale benchmarks by comparing the rankings produced by our aggregation algorithm to the leaderboard rankings. We achieve strong alignment with the ground truth rankings (Table \ref{tab:gt_results}).

\noindent \textcolor{kleinblue2}{\textbf{Q2. Is it better than current metrics?}} In addition to evaluating fit, we also compare our method to popular algorithms like Elo and LMArena. Table \ref{tab:gt_results} shows that our algorithm consistently outperforms these methods, demonstrating its superior performance for large real-world datasets.

\noindent \textcolor{kleinblue2}{\textbf{Q3. Are the top-k models preserved?}} 
A key concern for practitioners is whether the top models are ranked correctly. As shown in \cref{fig:rank_change}, our algorithm reliably preserves the ground-truth top-10 rankings, while other aggregation methods show weak and inconsistent alignment.

\noindent \textcolor{kleinblue2}{\textbf{Conclusion.}} The Plackett-Luce model fits real-world data well, outperforming other methods in both overall Kendall rank correlation and top-10 rankings, proving its effectiveness for large-scale benchmarks. The underlying reason is that we avoid using Elo distributions, which rely on assumptions that do not apply to foundation models \citep{boubdir2023elo}.

\subsubsection{Sample Efficiency and Robustness}
\begin{figure*}[htbp]
    \centering
    \begin{subfigure}{\textwidth}
        \centering
        \includegraphics[width=\textwidth]{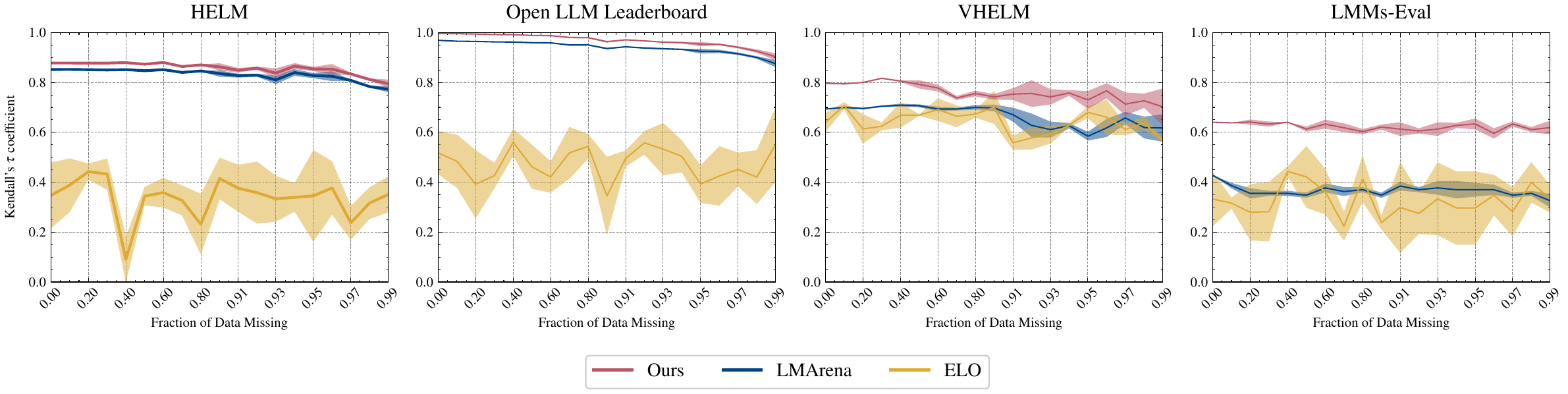}
        \label{subfig:sample-efficiency}
    \end{subfigure} \\
    \vspace{-5pt}
    \begin{subfigure}{\textwidth}
        \centering
        \includegraphics[width=\textwidth]{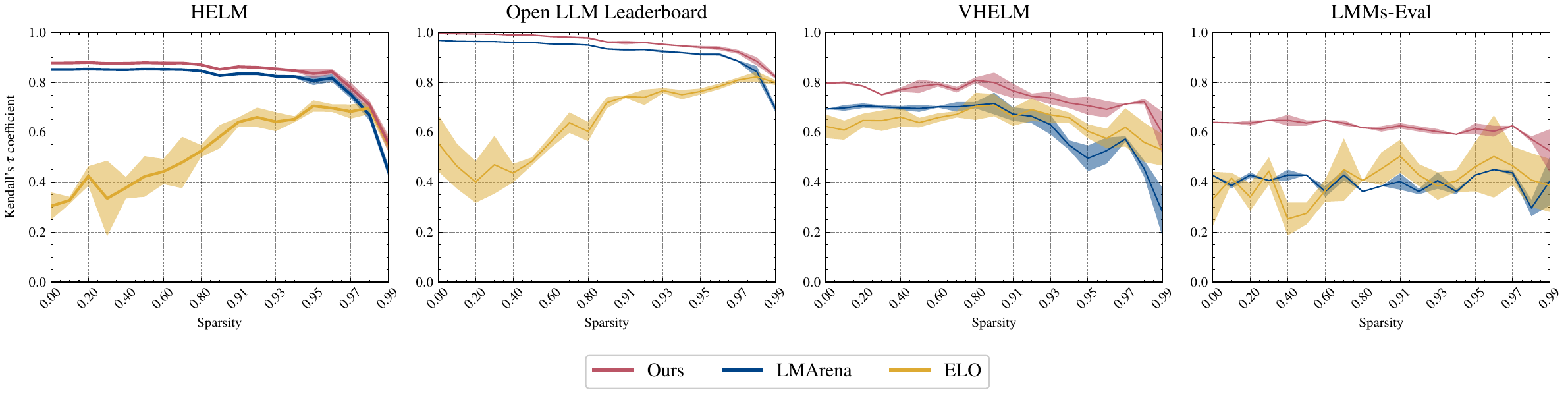}
        \label{subfig:sparsity}
    \end{subfigure}
    \vspace{-15pt}
    \caption{\textbf{Sample-efficient convergence and robustness to sparsity.} Kendall rank correlation coefficient between ground-truth ranking and different ranking methods as we randomly remove individual data samples (top) and model measurements (bottom). Methods typically remain robust to missing data, with Plackett-Luce consistently achieving higher correlation, even with 95\% measurements missing. The strongly convex Plackett-Luce model reliably converges to the same ranking.}
    \label{fig:robustness}
\end{figure*}

\noindent \textcolor{kleinblue2}{\textbf{Q1. Is Our Algorithm Sample-Efficient?}} We systematically reduce the number of data samples and re-rank the models using various methods, calculating Kendall’s $\tau$ for each. Missing data is simulated from 0\% to 99\%, with 10\% intervals until 90\%, followed by 1\% increments. As shown in \cref{fig:robustness}, our method maintains stable performance even with up to 95\% samples missing, demonstrating that it can achieve accurate rankings with up to 20 times fewer data points than current benchmarks.

\noindent \textcolor{kleinblue2}{\textbf{Q2. Can our Algorithm Aggregate Highly Sparse Rankings?}} We assess our method's ability to handle incomplete data by randomly removing a fraction of model measurements from each sample and re-ranking using the three aggregation methods.  We simulate data removal from 0\% to 99\%, with increments as before. As shown in \cref{fig:robustness}, our method remains effective even when 95\% of model comparisons are missing, proving it can recover accurate rankings with highly sparse data. This is crucial for ONEBench, which is designed to continually expand to accommodate more models and datasets, where models cannot be expected to be evaluated on the entire data pool.

\noindent \textcolor{kleinblue2}{\textbf{Conclusion.}} 
Our method is sample efficient and robust to sparse input rankings, maintaining accurate rankings with 20 times fewer data points.

\subsubsection{Comparison to other Social Choice Theory Frameworks}
Following the evaluation protocol used in ~\citet{rofin2022vote}, we compare our rank aggregation method with the Borda Count and the Dowdall system to provide a broader perspective on its performance against other social choice theory methods. As shown in \cref{tab:social_choice}, our method generally outperforms all the social choice theory frameworks, with drastic boosts on the more heterogeneous and incomplete benchmarks (VHELM and LMMs-Eval).  We attribute this to the fact that the Plackett-Luce method is a probabilistic scoring model that explicitly considers the uncertainty in rankings and captures inter-model agreement patterns, while Borda Count (by amplifying noise by assigning uniform linear weights to ranks) and the Dowdall System are both deterministic point-based methods. 
\label{subsec:social_choice}
\begin{table}[htbp]
\centering
\small 
\begin{tabular}{lccc}
\toprule
\textbf{Dataset} & \textbf{Borda} & \textbf{Dowdall} & \textbf{Ours} \\
\midrule
HELM        & 0.81 ± 0.00 & 0.83 ± 0.00 & \textbf{0.88 ± 0.00} \\
Leaderboard & 0.95 ± 0.00 & \textbf{0.99 ± 0.00}& \textbf{0.99 ± 0.00} \\
VHELM       & 0.35 ± 0.00 & 0.21 ± 0.00 & \textbf{0.79 ± 0.00} \\
LMMs-Eval   &  0.08 ± 0.00 & 0.18 ± 0.00 & \textbf{0.64 ± 0.00} \\
\bottomrule
\end{tabular}
\caption{\textbf{Comparison to other social choice theory methods.} Kendall rank correlations to ground-truth ranking across individual benchmarks.}
\label{tab:social_choice}
\end{table}

\section{ONEBench: Creation \& Capability Probing}

\label{infbench_sec}
\subsection{ONEBench-LLM}

\noindent \textcolor{kleinred}{\textbf{Data Pool $\mathcal{D}$.}} For ONEBench-LLM, we source data from the Open LLM Leaderboard, HELM, and LMArena. Open LLM Leaderboard and HELM aggregate several individual benchmarks, such as MMLU~\citep{hendrycks2020measuring} and HellaSwag~\citep{zellers2019hellaswag}, while LMArena uses pairwise model comparisons based on user-generated prompts.
Metrics include F1 score, exact match (EM), and quasi-exact match (QEM), as well as pairwise preferences.

\noindent \textcolor{kleinred}{\textbf{Models $\mathcal{M}$.}} For ONEBench-LLM, we use the 100 most downloaded models from Open LLM Leaderboard and all 79 models from HELM (as of v1.9.0), including both proprietary models like \texttt{GPT-4o} \citep{OpenAI2024} and open-weights ones like \texttt{LLaMA-3} \citep{Meta2024}. 

\subsection{ONEBench-LMM}

\noindent \textcolor{kleinred}{\textbf{Data Pool $\mathcal{D}$.}} For ONEBench-LMM, data is sourced from VHELM, LMMs-Eval, and WildVisionArena. Similar to ONEBench-LLM, VHELM and LMMs-Eval aggregate individual datasets like MMMU~\citep{yue2024mmmu} and VQAv2~\citep{goyal2017making}, while WildVisionArena uses pairwise tests for LMMs through image-based chats. 
Measurements include binary metrics like EM, QEM, and real-valued scores like ROUGE~\citep{lin2004rouge}.
We augment pairwise comparisons from WildVisionArena with LLM-as-a-Judge preferences generated using Prometheus-2~\citep{kim2024prometheus}, which correlate highly with human judgments. 

\noindent \textcolor{kleinred}{\textbf{Models $\mathcal{M}$.}} For ONEBench-LMM, we use 14 models from LMMs-Eval and 25 models from VHELM, including proprietary models like \texttt{Gemini Pro Vision}~\citep{team2023gemini} and open-weights models like \texttt{LLaVA}~\citep{liu2023llava}.

\subsection{Capability Probing}
\label{subsec:probing}

Given a query, the system retrieves relevant data samples using a combination of semantic and metadata search. This \textit{capability probing} provides a personalized comparison of foundation models.

\textcolor{kleinblue2}{\textbf{(i) Semantic search.}} We perform a nearest neighbours search in the embedding space of \texttt{all-MiniLM-L6-v2}~\citep{reimers-2019-sentence-bert} for language tasks and \texttt{SigLIP-B16}~\citep{zhai2023sigmoid} for vision-language tasks, using cosine similarity. We retrieve top $k$ samples for a given concept with tuned thresholds of $0.3$ for ONEBench-LLM and $0.7$ for ONEBench-LMM.

\textcolor{kleinblue2}{\textbf{(ii) Metadata search.}} We verify that per-sample metadata satisfies the constraints defined in the query. Some benchmarks, such as MMMU, are equipped with detailed metadata, including categories like image type, question type, area, etc.

\begin{figure*}[ht!]
    \centering
    \includegraphics[width=\textwidth]{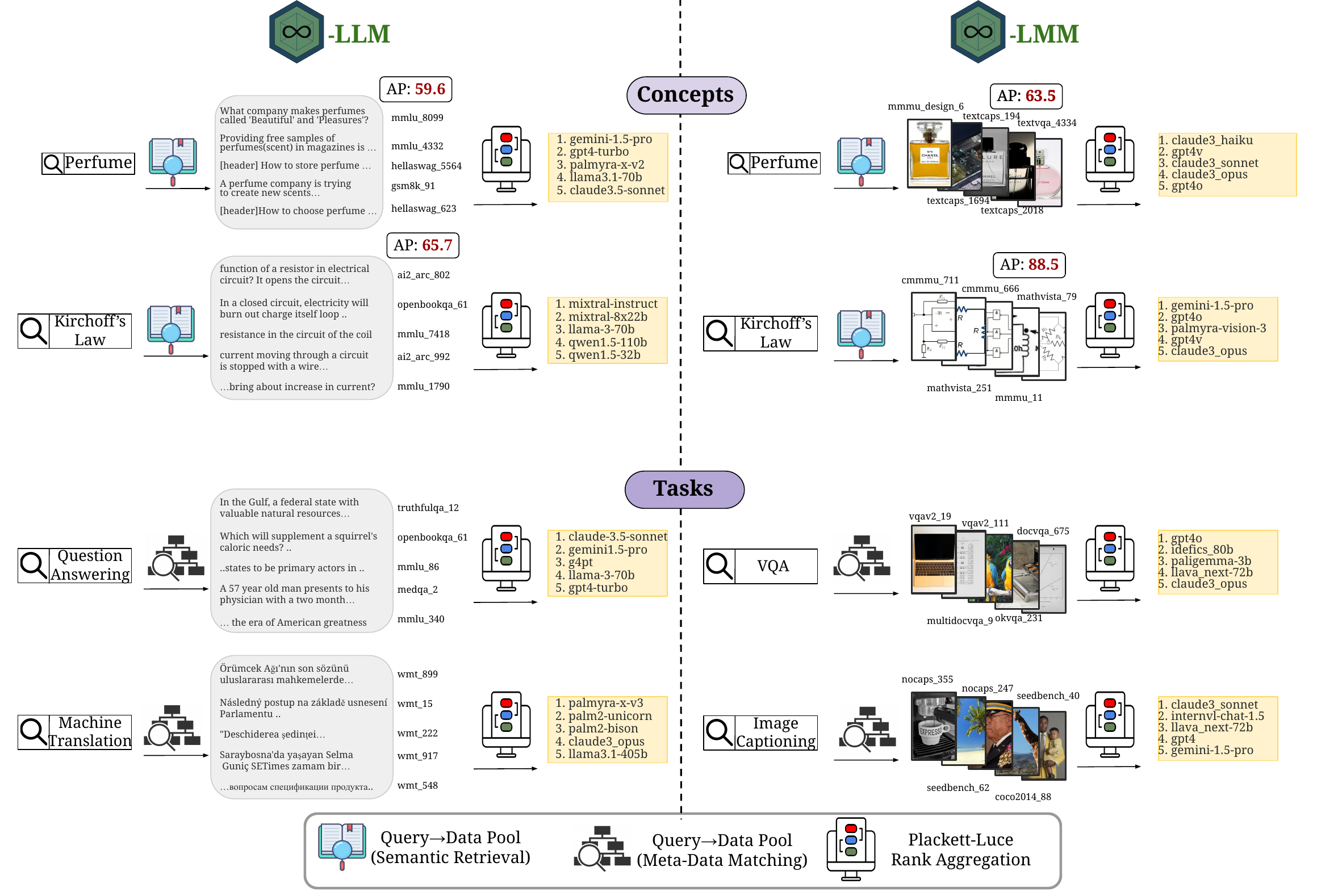}
    \caption{\textbf{Capability Probing (Qualitative):} we provide six sample retrieval results for a set of queries covering a diverse set of topics and report the top-5 models for each query.} 
    \label{fig:capability_probing}
\end{figure*}

Using these search mechanisms, we retrieve relevant samples from the data pool and aggregate ordinal model rankings with the Plackett-Luce model. We test ONEBench with a curated set of 50 concepts ranging from domain-specific knowledge, such as the Coriolis effect, to broader academic disciplines like neuroscience, and objects like the iPad (\cref{fig:capability_probing} and~\cref{appx:capability_probing}).

\begin{table}[htbp]
\centering
\resizebox{\linewidth}{!}{%
\begin{tabularx}{\linewidth}{l>{\raggedleft\arraybackslash}X>{\raggedleft\arraybackslash}X} 
    \toprule
    \textbf{Metric} & \textbf{LLM} & \textbf{LMM} \\
    \midrule
    Number of concepts & 40 & 50 \\
    Cohen-$\kappa$ & 0.79 & 0.91 \\
    mAP & 0.85 & 0.73 \\
    CMC@1 & 0.95 & 0.94 \\
    CMC@10 & 1.00 & 0.96 \\
    \bottomrule
\end{tabularx}
}
\caption{\textbf{Capability Probing (Quantitative)}: summary of accuracy and retrieval metrics.}
\label{tab:capability_probing}
\end{table}

\textcolor{kleinblue2}{\textbf{Insight 1. Retrieved samples are accurate and well-aligned with target concepts.}} Manual filtering out of incorrect samples by expert annotators\footnote{Inter-annotator agreement in~\cref{tab:capability_probing} shows strong consistency between annotators.  Annotators volunteered and provided informed consent. IRB approval was not obtained.} confirms high retrieval precision, with a mean Average Precision (mAP) of 0.85 (LLM) and 0.73 (LMM). These results, shown in~\cref{tab:capability_probing}, indicate that our system reliably retrieves samples matching the intended capabilities, with room for improvement on some underrepresented concepts. Please refer to the per-concept AP in \cref{tab:ap_comparison} for a better indicator.
Note that the retrieval mechanism is expected to only improve with better models and larger test sets covering more diverse capabilities.

\textcolor{kleinblue2}{\textbf{Insight 2. Top models vary significantly across queries.}} A key aspect of evaluating capability-specific performance is checking whether the set of top-performing models remains consistent across different queries. If rankings were largely stable, with the same models dominating regardless of the capability assessed, fine-grained querying would offer limited practical value. However, our results in \cref{fig:capability_probing} and \cref{fig:capability_probing_appendix} reveal substantial variation in the top-$k$ models across domains and concepts. This variation highlights that relying solely on global leaderboards can obscure meaningful differences. These findings confirm that ONEBench enables more targeted model selection by surfacing the most appropriate candidates for a given capability query.
\begin{table}[htbp]
\centering
\begin{tabular}{lcc}
\toprule
\textbf{Samples Retrieved} & \textbf{Random $\tau$} & \textbf{Query $\tau$} \\
\midrule
100 & 0.78 ± 0.04 & 0.68 ± 0.10 \\
1,000 & 0.93 ± 0.01 & 0.83 ± 0.05 \\
10,000 & 0.98 ± 0.00 & 0.91 ± 0.03 \\
\bottomrule
\end{tabular}
\caption{\textbf{Capability probing vs. random retrieval:} Average Kendall's $\tau$ with the global ranking.}
\label{tab:query_vs_random}
\end{table}

\textcolor{kleinblue2}{\textbf{Insight 3. More specific queries lead to more distinct model rankings.}}  
We conduct an experiment analyzing the divergence between global and capability-specific model orderings to assess how domain specificity affects ranking. Similar rankings would suggest that domain-specific evaluation adds limited value. For each of the 40 concepts in~\cref{appx:capability_probing}, we retrieve $n$ relevant samples and compute the average Kendall rank correlation with the global ranking. As a baseline, we perform five random retrievals of $n$ samples from the full pool and compute their average correlation. Results in~\cref{tab:query_vs_random} show that concept-specific rankings are consistently less correlated with the global ranking than random ones, even as $n$ increases. This suggests that models specialize across domains and should be evaluated accordingly.
\section{Related Works}
Recent multi-task benchmarks, such as GLUE~\citep{wang2018glue}, SuperGLUE~\citep{wang2019superglue}, and BigBench~\citep{srivastava2022beyond}, test the broad capabilities of foundation models. However, these benchmarks use arithmetic mean for task aggregation~\citep{open-llm-leaderboard} which can distort rankings \citep{zhang2024inherent} and is sensitive to outliers \citep{agarwal2021deep} or missing scores \citep{himmi2023towards}. ONEBench addresses these by enabling sample reuse, avoiding task selection bias \citep{dominguez2024training}.  Inspired by social choice theory, ONEBench employs ordinal rankings and the Plackett-Luce model for aggregation, which is robust to irrelevant alternatives and outliers. Moreover, ONEBench reduces evaluation costs, similar to compressed subsets \citep{zhao2024flasheval} and lifelong benchmarks \citep{prabhu2024lifelong}. Further, by flexibly integrating diverse sample and measurements contributions, ONEBench can be more inclusive than traditional benchmarks dominated by well-funded institutions~\citep{pouget2024no, nguyen2024multilingual}. We provide an expanded review in \cref{appx:rw}. 

\section{Conclusions and Open Problems}
\label{sec:conclusion}
We introduce ONEBench, an open-ended benchmarking framework for foundation models. Our open, democratized benchmarking methodology allows various stakeholders to contribute evaluation samples and model measurements with detailed metadata. This affords creating customized benchmarks and testing arbitrary capabilities with semantic and structured searches. We provide an aggregation mechanism that is both theoretically grounded and empirically validated to be robust to incomplete data and heterogeneous measurements across evaluations. We demonstrate the utility of ONEBench for LLMs and LMMs, showing how dynamic probing reveals new insights into model performance on specific tasks and concepts. This combination of theoretical rigour, empirical results, and practical flexibility makes ONEBench a valuable tool for comprehensive evaluation.

\section{Limitations}
\label{sec:limitations}

Our approach, while promising, comes with its share of challenges. We highlight three key issues.

\textcolor{kleinblue2}{\textbf{Effects of Aggregation.}} Combining different types of evaluation data into a single ranking risks oversimplifying important performance differences. We mitigate this by introducing flexible querying. Furthermore, conversion to pairwise ranking leads to loss of information which could hurt aggregation algorithms. However, in real-world scenarios pairwise measurements perform better, despite information loss \citep{shah2014better}.

\textcolor{kleinblue2}{\textbf{Reproducibility.}} The dynamic nature of capability querying and the expanding sample pool, though useful, makes it harder to maintain consistency and can introduce bias during data collection and aggregation. However, our framework supports consistent evaluation, as each sample has a unique identifier and the querying mechanism is deterministic. Reproducibility can be achieved simply by specifying which samples are included.

\textcolor{kleinblue2}{\textbf{Statistical Modeling Assumptions.}} Our reliance on statistical models like Plackett–Luce might make assumptions about data distribution that may not always hold, affecting the reliability of our results. This limitation is not unique to our approach, but holds for any method that makes modeling assumptions. \citet{noothigattu2020axioms} show that Plackett–Luce based models do not satisfy the following axioms:

\textcolor{kleinblue2}{\textbf{(i) Pairwise Majority Consistency.}} If pairwise preference order across models are consistent: a > b, b > c and a > c, then ranking should preserve the consistency: a > b > c.

\textcolor{kleinblue2}{\textbf{(ii) Separability.}} If model a is higher than model b in MLE estimate scores in two input sets, a must be higher than b in MLE estimate scores of their combined set. We show empirically that violating separability is not a problem in practice. We present results in~\cref{appx:separability}.

Overall, we believe democratic, open-ended benchmarking is an impactful direction to explore, despite the apparent limitations.

\section{Broad Impacts}

Our work could have a meaningful impact on efficacy of benchmarking for foundation models. With ONEBench, we offer a benchmarking framework that can adapt to different domains, allowing for more inclusive and transparent evaluation practices, empowering researchers and downstream practitioners. By making benchmarking more accessible, we hope to encourage fairness, reproducibility, and innovation in how evaluation frameworks are designed. In the long run, this approach can help build a deeper understanding of foundation models across both language and vision–language tasks. We do not believe that there are any immediate negative societal consequences as a result of this work, but caution that all findings are preliminary and need additional evaluation before deployment.

\section*{Acknowledgements}
The authors would like to thank (in alphabetic order): Nikhil Chandak, Karel D'Oosterlinck, Shashwat Goel, Shyamgopal Karthik, Bálint Mucsányi and Tim Weiland for their helpful suggestions and feedback. Special thanks to Bo Li for (promptly and consistently) providing sample-level model evaluation logs of LMMs-Eval, Yujie Lu for providing early access to the battle data of Vision Arena and Palzer Lama for creating the logo of ONEBench and helping with figures. 
The authors would also like to thank Federico D'Agostino for recommending the title of the paper and the name of the benchmark.
AP and MB acknowledge financial support by the Federal Ministry of Education and Research (BMBF), FKZ: 011524085B and Open Philanthropy Foundation funded by the Good Ventures Foundation.  VU and SD thank the International Max Planck Research School for Intelligent Systems (IMPRS-IS) and the European Laboratory for Learning and Intelligent Systems (ELLIS) PhD program for support. 
VU is supported by a Google PhD Fellowship in Machine Intelligence.
MB acknowledges support via the CRC 1233 on Robust Vision and from the Machine Learning Cluster of Excellence, funded by the Deutsche Forschungsgemeinschaft (DFG, German Research Foundation) under Germany’s Excellence Strategy – EXC number 2064/1 – Project number 390727645.

\bibliography{main}
\clearpage
\appendix
\onecolumn

\stoptocwriting\
\setcounter{page}{1}
\doparttoc 
\faketableofcontents 

\part{Appendix} 
\parttoc 
\clearpage

\section{Datasets used in ONEBench: Further Details}
\label{appx:datasets}
\begin{figure*}[htbp]
    \centering
    \includegraphics[width=\textwidth]{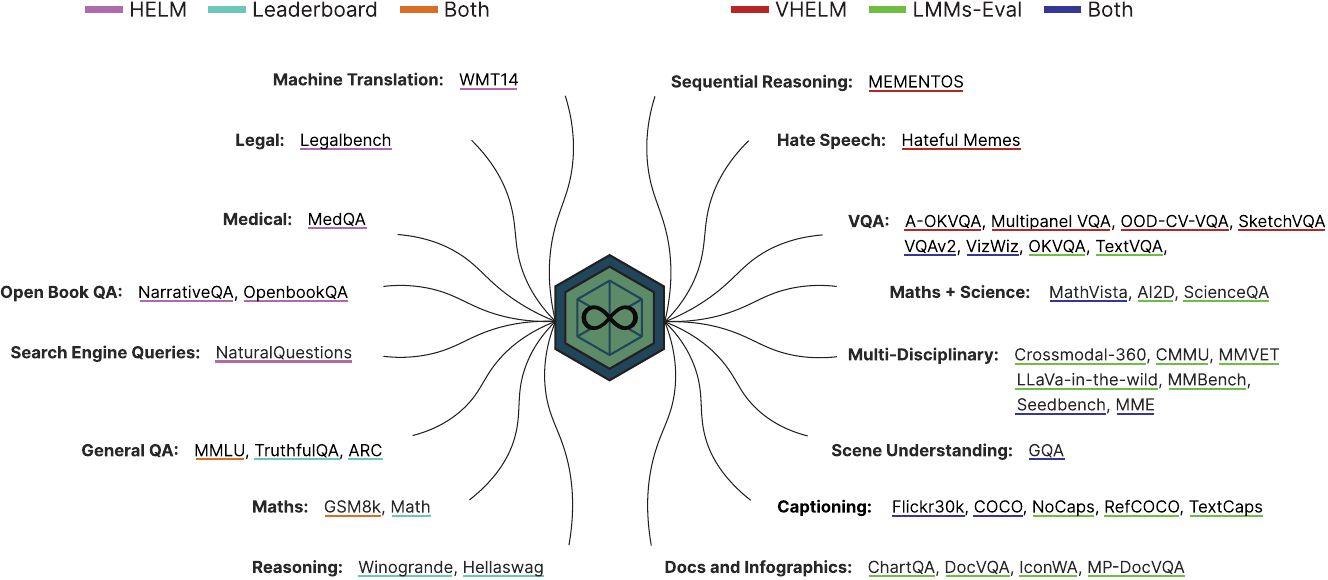}
    \caption{\textbf{Constituent datasets of ONEBench-LLM (left) and OneBench-LMM (right)}. We provide task type, metric, and license about each dataset in \cref{tab:llms-details} and \cref{tab:lmms-details}.}
    \label{fig:benchmarks}
\end{figure*}

\begin{table*}[htbp]
\centering
{\scriptsize
\resizebox{\textwidth}{!}{

\begin{tabular}{l|c|c|c|c|c}
\toprule

\textbf{Dataset} & \textbf{Source} & \textbf{Task} & \textbf{Size} & \textbf{Metric} & \textbf{License} \\
\midrule
\multicolumn{6}{c}{\textbf{Cardinal}}\\
\midrule

LegalBench~\citep{guha2024legalbench} & HELM & Legal & 1K & QEM &Unknown\\
MATH~\citep{hendrycksmath2021} & HELM & Maths & 1K & QEM & MIT\\
MedQA~\citep{jin2021disease} & HELM & Medical & 1K & QEM &MIT\\
NarrativeQA~\citep{kovcisky2018narrativeqa} & HELM & Openbook QA & 1K & F1 &Apache-2.0\\
NaturalQuestions~\citep{kwiatkowski2019natural} & HELM & Search Engine Queries & 1K & F1 &CC BY-SA 3.0\\
OpenbookQA~\citep{mihaylov2018can} & HELM & Openbook QA & 1K & EM &Apache-2.0 \\
WMT 2014~\citep{bojar2014findings} & HELM & Machine translation & 1K & BLEU &CC-BY-SA-4.0\\

ARC~\citep{clark2018think} & Leaderboard & General QA &$1.1$K& EM &CC-BY-SA-4.0\\
HellaSwag~\citep{zellers2019hellaswag} & Leaderboard & Reasoning &$10$K& EM &MIT\\
TruthfulQA~\citep{lin2022truthfulqa} & Leaderboard & General QA &$817$& EM &Apache-2.0\\
Winogrande~\citep{sakaguchi2021winogrande} & Leaderboard & Reasoning &$1.2$K& EM &Apache-2.0\\

GSM8K~\citep{cobbe2021training}  & HELM + Leaderboard & Maths &$1.3$K& QEM &MIT\\
MMLU~\citep{hendrycks2020measuring} & HELM + Leaderboard & General QA &$13.8$K& EM &MIT\\

\midrule
\multicolumn{6}{c}{\textbf{Ordinal}}\\
\midrule
Chatbot Arena~\cite{chiang2024chatbot} & Chatbot Arena & Pairwise Battles & $51$K & - &CC BY 4.0\\
\bottomrule
\end{tabular}}}
\caption{\textbf{Datasets in ONEBench-LLM}. A diverse collection of benchmarks testing the abilities of LLMs in areas such as law, medicine, mathematics, question answering, reasoning and instruction following, as well as the performance of LLMs in pairwise battles.}
\label{tab:llms-details}
\end{table*}

\begin{table*}[htbp]
\centering
{\scriptsize
\resizebox{\textwidth}{!}{%

\begin{tabular}{l|c|c|c|c|c}
\toprule

\textbf{Dataset} & \textbf{Source} & \textbf{Task} & \textbf{Size} & \textbf{Metric} & \textbf{License} \\
\midrule
\multicolumn{6}{c}{\textbf{Cardinal}}\\
\midrule
A-OKVQA~\citep{schwenk2022okvqa} & VHELM & VQA & $7.2$K& QEM & Apache-2.0\\
Bingo~\citep{cui2023holistic}  & VHELM & Bias+Hallucination &$886$ & ROUGE & Unknown\\
Crossmodal-3600~\citep{thapliyal2022crossmodal} & VHELM & Captioning &$1.5$K& ROUGE &CC BY-SA 4.0\\
Hateful Memes~\citep{kiela2020hateful}& VHELM & Hate Speech &$1$K& QEM& Custom(Meta) \\
Mementos~\citep{wang2024mementos} & VHELM & Sequential Reasoning &$945$& GPT & CC-BY-SA-4.0\\
MultipanelVQA~\citep{fan2024muffin} & VHELM & VQA &$200$& QEM& MIT\\
OODCV-VQA~\citep{tu2023how}&VHELM & VQA &$1$K& QEM& CC-BY-NC-4.0\\
PAIRS~\citep{fraser-kiritchenko-2024-examining}& VHELM & Bias &$508$& QEM & Unknown\\
Sketchy-VQA~\citep{tu2023how}& VHELM & VQA &$1$K& QEM &CC-BY-NC-4.0\\
AI2D~\citep{kembhavi2016diagram} & LMMs-Eval & Maths+Science &$3.09$K& QEM & Apache-2.0\\
IconQA~\citep{lu2021iconqa} & LMMs-Eval & Docs and Infographics &$43$K& ANLS &CC BY-SA 4.0\\
InfoVQA~\citep{mathew2022infographicvqa} & LMMs-Eval & Docs and Infographics &$6.1$K& ANLS &Unknown\\
LLaVA-in-the-Wild~\citep{liu2023llava} & LMMs-Eval & Multi-disciplinary &$60$& GPT4 &Apache-2.0\\
ChartQA~\citep{masry2022chartqa} & LMMs-Eval & Docs and Infographics &$2.5$K& QEM & GPL-3.0\\
CMMMU~\cite{zhang2024cmmmu}& LMMs-Eval & Multi-disciplinary &$900$& QEM &CC-BY-4.0\\
DocVQA~\citep{mathew2021docvqa} & LMMs-Eval & Docs and Infographics &$10.5$K& ANLS &Unknown\\
MMBench~\citep{liu2023mmbench} & LMMs-Eval & Multi-disciplinary &$24$K& GPT & Apache-2.0\\
MMVET~\citep{yu2024mmvet} & LMMs-Eval & Multi-disciplinary &$218$& GPT & Apache-2.0\\
MP-DocVQA~\citep{tito2023hierarchical} & LMMs-Eval & Docs and Infographics &$5.2$K& QEM &MIT\\
NoCaps~\citep{agrawal2019nocaps} & LMMs-Eval & Captioning &$4.5$K& ROUGE&MIT\\
OK-VQA~\citep{marino2019ok} & LMMs-Eval & VQA &$5.1$K& ANLS & Unknown\\
RefCOCO~\citep{kazemzadeh2014referitgame,mao2016generation} & LMMs-Eval & Captioning &$38$K& ROUGE & Apache-2.0\\
ScienceQA~\citep{lu2022learn} & LMMs-Eval & Maths+Science &$12.6$K& EM &CC BY-NC-SA 4.0\\
TextCaps~\citep{sidorov2020textcaps} & LMMs-Eval & Captioning &$3.2$K& ROUGE&CC BY 4.0\\
TextVQA~\citep{singh2019towards} & LMMs-Eval & VQA &$5$K& EM &CC BY 4.0\\
COCO~\citep{lin2014microsoft} & VHELM+LMMs-Eval & Captioning &$45.5$K& ROUGE & CC-BY-4.0\\
Flickr30k~\citep{young2014image} & VHELM+LMMs-Eval & Captioning &$31$K& ROUGE &CC-0 Public Domain\\
GQA\citep{hudson2019gqa} & VHELM+LMMs-Eval & Scene Understanding &$12.6$K& QEM &CC-BY-4.0\\
MathVista~\citep{lu2024mathvista} & VHELM+LMMs-Eval & Maths+Science &$1$K& QEM/GPT4& CC-BY-SA-4.0\\
MME~\citep{fu2023mme} & VHELM+LMMs-Eval & Multi-disciplinary &$2.4$K& QEM/C+P & Unknown\\
MMMU~\citep{yue2024mmmu} & VHELM+LMMs-Eval & Multi-disciplinary &$900$& QEM& CC BY-SA 4.0\\
POPE~\citep{li2023evaluating} & VHELM+LMMs-Eval & Hallucination &$9$K& QEM/EM & MIT\\
SEED-Bench~\citep{li2023seed,li2024seed} &VHELM+LMMs-Eval & Multi-disciplinary &$42.5$K& QEM/EM & Apache\\
VizWiz~\citep{gurari2018vizwiz} & VHELM+LMMs-Eval & VQA &$4.3$K& QEM/EM &CC BY 4.0\\
VQAv2~\citep{goyal2017making} & VHELM+LMMs-Eval & VQA &$214$K& QEM/EM &CC BY 4.0\\
\midrule
\multicolumn{6}{c}{\textbf{Ordinal}}\\
\midrule
Vision Arena~\citep{yujie2024wildvisionarena} & - & Pairwise Battles & $9$K & - & MIT\\
LMMs-Eval(Prometheus2)~\citep{kim2024prometheus} & - & Pairwise Battles & $610$K & - & MIT \\
\bottomrule
\end{tabular}}}
\caption{\textbf{Datasets in ONEBench-LMM}: a diverse collection of benchmarks testing the abilities of LLMs in tasks such as general VQA, image captioning, hate speech detection, bias and hallucination understanding, maths and science, documents and infographics, scene understanding and sequential reasoning as well as the performance of LMMs in pairwise battles. Additional preference comparisons are sampled randomly from LMMs-Eval, which are excluded from the  cardinal measurement sample pool.}
\label{tab:lmms-details}
\end{table*}

\pagebreak
\clearpage
\section{Models used in ONEBench: Further Details}
\label{appx:models}
In this section, we provide a deeper insight into the models used in the creation of ONEBench. It is important to note that ONEBench-LLM and ONEBench-LMM have complementary characteristics: while ONEBench-LLM has fewer data samples $\mathcal{D}_k$, they are evaluated on more models $\mathcal{M}_k$, while ONEBench-LMM contains (significantly) more data samples but they are evaluated on less models. 

\subsection{ONEBench-LLM: Open LLM Leaderboard}
The Open LLM Leaderboard~\citep{open-llm-leaderboard} was created to track progress of LLMs in the open-source community by evaluating models on the same data samples and setup for more reproducible results and a trustworthy leaderboard where all open-sourced LLMs could be ranked.\vspace{0.15cm}

However, due to the abundance of models found on the leaderboard and the lack of adequate documentation, and therefore reliability, of many of these models being evaluated, we rank the models based on the number of downloads, as a metric of adoption of these models by the community. We provide the total list of models as an artefact and list the top 100 models below:

{\setlength{\columnsep}{3pt} 
\small
\begin{enumerate}
\item \texttt{01-ai/Yi-34B-200K}
\item \texttt{AI-Sweden-Models/gpt-sw3-126m}
\item \texttt{BioMistral/BioMistral-7B}
\item \texttt{CohereForAI/c4ai-command-r-plus}
\item \texttt{CohereForAI/c4ai-command-r-v01}
\item \texttt{Deci/DeciLM-7B-instruct}
\item \texttt{EleutherAI/llemma$\_$7b}
\item \texttt{EleutherAI/pythia-410m}
\item \texttt{Felladrin/Llama-160M-Chat-v1}
\item \texttt{Felladrin/Llama-68M-Chat-v1}
\item \texttt{FreedomIntelligence/AceGPT-7B}
\item \texttt{GritLM/GritLM-7B}
\item \texttt{Intel/neural-chat-7b-v3-1}
\item \texttt{JackFram/llama-160m}
\item \texttt{Nexusflow/NexusRaven-V2-13B}
\item \texttt{Nexusflow/Starling-LM-7B-beta}
\item \texttt{NousResearch/Hermes-2-Pro-Mistral-7B}
\item \texttt{NousResearch/Meta-Llama-3-8B-Instruct}
\item \texttt{NousResearch/Nous-Hermes-2-Mixtral-8x7B-DPO}
\item \texttt{NousResearch/Nous-Hermes-2-SOLAR-10.7B}
\item \texttt{NousResearch/Nous-Hermes-2-Yi-34B}
\item \texttt{OpenPipe/mistral-ft-optimized-1227}
\item \texttt{Qwen/Qwen1.5-0.5B}
\item \texttt{Qwen/Qwen1.5-0.5B-Chat}
\item \texttt{Qwen/Qwen1.5-1.8B}
\item \texttt{Qwen/Qwen1.5-1.8B-Chat}
\item \texttt{Qwen/Qwen1.5-110B-Chat}
\item \texttt{Qwen/Qwen1.5-14B}
\item \texttt{Qwen/Qwen1.5-14B-Chat}
\item \texttt{Qwen/Qwen1.5-32B-Chat}
\item \texttt{Qwen/Qwen1.5-4B}
\item \texttt{Qwen/Qwen1.5-4B-Chat}
\item \texttt{Qwen/Qwen1.5-72B-Chat}
\item \texttt{Qwen/Qwen1.5-7B}
\item \texttt{Qwen/Qwen1.5-7B-Chat}
\item \texttt{SeaLLMs/SeaLLM-7B-v2}
\item \texttt{TinyLlama/TinyLlama-1.1B-Chat-v1.0}
\item \texttt{TinyLlama/TinyLlama-1.1B-intermediate-step-3T}
\item \texttt{VAGOsolutions/SauerkrautLM-Mixtral-8x7B}
\item \texttt{abhishekchohan/mistral-7B-forest-dpo}
\item \texttt{ahxt/LiteLlama-460M-1T}
\item \texttt{ai-forever/mGPT}
\item \texttt{alignment-handbook/zephyr-7b-sft-full}
\item \texttt{augmxnt/shisa-gamma-7b-v1}
\item \texttt{bigcode/starcoder2-15b}
\item \texttt{bigcode/starcoder2-3b}
\item \texttt{bigcode/starcoder2-7b}
\item \texttt{cloudyu/Mixtral$\_$7Bx4$\_$MOE$\_$24B}
\item \texttt{codellama/CodeLlama-70b-Instruct-hf}
\item \texttt{cognitivecomputations/dolphin-2.2.1-mistral-7b}
\item \texttt{cognitivecomputations/dolphin-2.6-mistral-7b-dpo}
\item \texttt{cognitivecomputations/dolphin-2.9-llama3-8b}
\item \texttt{daekeun-ml/phi-2-ko-v0.1}
\item \texttt{deepseek-ai/deepseek-coder-1.3b-instruct}
\item \texttt{deepseek-ai/deepseek-coder-6.7b-base}
\item \texttt{deepseek-ai/deepseek-coder-6.7b-instruct}
\item \texttt{deepseek-ai/deepseek-coder-7b-instruct-v1.5}
\item \texttt{deepseek-ai/deepseek-math-7b-base}
\item \texttt{deepseek-ai/deepseek-math-7b-instruct}
\item \texttt{deepseek-ai/deepseek-math-7b-rl}
\item \texttt{google/codegemma-7b-it}
\item \texttt{google/gemma-1.1-7b-it}
\item \texttt{google/gemma-2b}
\item \texttt{google/gemma-2b-it}
\item \texttt{google/gemma-7b}
\item \texttt{google/gemma-7b-it}
\item \texttt{google/recurrentgemma-2b-it}
\item \texttt{h2oai/h2o-danube2-1.8b-chat}
\item \texttt{hfl/chinese-alpaca-2-13b}
\item \texttt{ibm/merlinite-7b}
\item \texttt{meta-llama/Meta-Llama-3-70B}
\item \texttt{meta-llama/Meta-Llama-3-70B-Instruct}
\item \texttt{meta-llama/Meta-Llama-3-8B}
\item \texttt{meta-llama/Meta-Llama-3-8B-Instruct}
\item \texttt{meta-math/MetaMath-Mistral-7B}
\item \texttt{microsoft/Orca-2-7b}
\item \texttt{microsoft/phi-2}
\item \texttt{mistral-community/Mistral-7B-v0.2}
\item \texttt{mistral-community/Mixtral-8x22B-v0.1}
\item \texttt{mistralai/Mistral-7B-Instruct-v0.2}
\item \texttt{mistralai/Mixtral-8x22B-Instruct-v0.1}
\item \texttt{mistralai/Mixtral-8x7B-Instruct-v0.1}
\item \texttt{mistralai/Mixtral-8x7B-v0.1}
\item \texttt{openai-community/gpt2}
\item \texttt{openai-community/gpt2-large}
\item \texttt{openchat/openchat-3.5-0106}
\item \texttt{openchat/openchat-3.5-1210}
\item \texttt{openchat/openchat$\_$3.5}
\item \texttt{sarvamai/OpenHathi-7B-Hi-v0.1-Base}
\item \texttt{speakleash/Bielik-7B-Instruct-v0.1}
\item \texttt{speakleash/Bielik-7B-v0.1}
\item \texttt{stabilityai/stablelm-2-1$\_$6b}
\item \texttt{stabilityai/stablelm-2-zephyr-1$\_$6b}
\item \texttt{stabilityai/stablelm-zephyr-3b}
\item \texttt{teknium/OpenHermes-2.5-Mistral-7B}
\item \texttt{tokyotech-llm/Swallow-70b-instruct-hf}
\item \texttt{upstage/SOLAR-10.7B-Instruct-v1.0}
\item \texttt{upstage/SOLAR-10.7B-v1.0}
\item \texttt{wenbopan/Faro-Yi-9B}
\item \texttt{yanolja/EEVE-Korean-Instruct-10.8B-v1.0}
\end{enumerate}
}

\subsection{ONEBench-LLM: HELM}
Similar to the Open LLM Leaderboard, the goal of HELM was to provide a uniform evaluation of language models over a  vast set of data samples (termed as \texttt{scenarios} in \citet{liang2022holistic}). HELM, however, has a broader scope of models used for evaluation, employing open, limited-access, and closed models. All models currently used in ONEBench-LLM is listed below:

{\setlength{\columnsep}{3pt} 
\small
\begin{enumerate}
\item \texttt{01-ai$\_$yi-34b}
\item \texttt{01-ai$\_$yi-6b}
\item \texttt{01-ai$\_$yi-large-preview}
\item \texttt{ai21$\_$j2-grande}
\item \texttt{ai21$\_$j2-jumbo}
\item \texttt{ai21$\_$jamba-1.5-large}
\item \texttt{ai21$\_$jamba-1.5-mini}
\item \texttt{ai21$\_$jamba-instruct}
\item \texttt{AlephAlpha$\_$luminous-base}
\item \texttt{AlephAlpha$\_$luminous-extended}
\item \texttt{AlephAlpha$\_$luminous-supreme}
\item \texttt{allenai$\_$olmo-7b}
\item \texttt{anthropic$\_$claude-2.0}
\item \texttt{anthropic$\_$claude-2.1}
\item \texttt{anthropic$\_$claude-3-5-sonnet-20240620}
\item \texttt{anthropic$\_$claude-3-haiku-20240307}
\item \texttt{anthropic$\_$claude-3-opus-20240229}
\item \texttt{anthropic$\_$claude-3-sonnet-20240229}
\item \texttt{anthropic$\_$claude-instant-1.2}
\item \texttt{anthropic$\_$claude-instant-v1}
\item \texttt{anthropic$\_$claude-v1.3}
\item \texttt{cohere$\_$command}
\item \texttt{cohere$\_$command-light}
\item \texttt{cohere$\_$command-r}
\item \texttt{cohere$\_$command-r-plus}
\item \texttt{databricks$\_$dbrx-instruct}
\item \texttt{deepseek-ai$\_$deepseek-llm-67b-chat}
\item \texttt{google$\_$gemini-1.0-pro-001}
\item \texttt{google$\_$gemini-1.0-pro-002}
\item \texttt{google$\_$gemini-1.5-flash-001}
\item \texttt{google$\_$gemini-1.5-pro-001} 
\item \texttt{google$\_$gemini-1.5-pro-preview-0409}
\item \texttt{google$\_$gemma-2-9b-it}
\item \texttt{google$\_$gemma-2-27b-it}
\item \texttt{google$\_$gemma-7b}
\item \texttt{google$\_$text-bison@001}
\item \texttt{google$\_$text-unicorn@001}
\item \texttt{meta$\_$llama-2-7b}
\item \texttt{meta$\_$llama-2-13b}
\item \texttt{meta$\_$llama-2-70b}
\item \texttt{meta$\_$llama-3-8b}
\item \texttt{meta$\_$llama-3-70b}
\item \texttt{meta$\_$llama-3.1-8b-instruct-turbo}
\item \texttt{meta$\_$llama-3.1-70b-instruct-turbo}
\item \texttt{meta$\_$llama-3.1-405b-instruct-turbo}
\item \texttt{meta$\_$llama-65b}
\item \texttt{microsoft$\_$phi-2}
\item \texttt{microsoft$\_$phi-3-medium-4k-instruct}
\item \texttt{microsoft$\_$phi-3-small-8k-instruct}
\item \texttt{mistralai$\_$mistral-7b-instruct-v0.3}
\item \texttt{mistralai$\_$mistral-7b-v0.1}
\item \texttt{mistralai$\_$mistral-large-2402}
\item \texttt{mistralai$\_$mistral-large-2407}
\item \texttt{mistralai$\_$mistral-medium-2312}
\item \texttt{mistralai$\_$mistral-small-2402}
\item \texttt{mistralai$\_$mixtral-8x7b-32kseqlen}
\item \texttt{mistralai$\_$mixtral-8x22b}
\item \texttt{mistralai$\_$open-mistral-nemo-2407}
\item \texttt{nvidia$\_$nemotron-4-340b-instruct}
\item \texttt{openai$\_$gpt-3.5-turbo-0613}
\item \texttt{openai$\_$gpt-4-0613}
\item \texttt{openai$\_$gpt-4-1106-preview}
\item \texttt{openai$\_$gpt-4-turbo-2024-04-09}
\item \texttt{openai$\_$gpt-4o-2024-05-13}
\item \texttt{openai$\_$gpt-4o-mini-2024-07-18}
\item \texttt{openai$\_$text-davinci-002}
\item \texttt{openai$\_$text-davinci-003}
\item \texttt{qwen$\_$qwen1.5-7b}
\item \texttt{qwen$\_$qwen1.5-14b}
\item \texttt{qwen$\_$qwen1.5-32b}
\item \texttt{qwen$\_$qwen1.5-72b}
\item \texttt{qwen$\_$qwen1.5-110b-chat}
\item \texttt{qwen$\_$qwen2-72b-instruct}
\item \texttt{snowflake$\_$snowflake-arctic-instruct}
\item \texttt{tiiuae$\_$falcon-7b}
\item \texttt{tiiuae$\_$falcon-40b}
\item \texttt{writer$\_$palmyra-x-004}
\item \texttt{writer$\_$palmyra-x-v2}
\item \texttt{writer$\_$palmyra-x-v3}
\end{enumerate}
}

\subsection{ONEBench-LMM: LMMs-Eval}
LMMs-Eval is the first comprehensive large-scale evaluation benchmark for Large Multimodal models, meant ``to promote transparent and reproducible evaluations''~\citep{zhang2024lmms}. The models supported by LMMs-Eval are primarily open-sourced and the full list of currently used models are listed below:
{\setlength{\columnsep}{3pt} 
\small
\begin{enumerate}
\item \texttt{idefics2-8b}
\item \texttt{internlm-xcomposer2-4khd-7b}
\item \texttt{instructblip-vicuna-7b}
\item \texttt{instructblip-vicuna-13b}
\item \texttt{internVL-Chat-V1-5}
\item \texttt{llava-13b}
\item \texttt{llava-1.6-13b}
\item \texttt{llava-1.6-34b}
\item \texttt{llava-1.6-mistral-7b}
\item \texttt{llava-1.6-vicuna-13b}
\item \texttt{llava-1.6-vicuna-7b}
\item \texttt{llava-7b}
\item \texttt{llava-next-72b}
\item \texttt{qwen$\_$vl$\_$chat}
\end{enumerate}
}

\subsection{ONEBench-LMM: VHELM}
Finally, ONEBench-LMM comprises VHELM, an extension of HELM for Vision-Language models. The models currently used by us, spanning open, limited-access, and closed models, are as follows:

{\setlength{\columnsep}{3pt} 
\small
\begin{enumerate}
\item \texttt{anthropic$\_$claude$\_$3$\_$haiku$\_$20240307}
\item \texttt{anthropic$\_$claude$\_$3$\_$opus$\_$20240229}
\item \texttt{anthropic$\_$claude$\_$3$\_$sonnet$\_$20240229}
\item \texttt{google$\_$gemini$\_$1.0$\_$pro$\_$vision$\_$001}
\item \texttt{google$\_$gemini$\_$1.5$\_$pro$\_$preview$\_$0409}
\item \texttt{google$\_$gemini$\_$pro$\_$vision}
\item \texttt{google$\_$paligemma$\_$3b$\_$mix$\_$448}
\item \texttt{huggingfacem4$\_$idefics2$\_$8b}
\item \texttt{huggingfacem4$\_$idefics$\_$80b}
\item \texttt{huggingfacem4$\_$idefics$\_$80b$\_$instruct}
\item \texttt{huggingfacem4$\_$idefics$\_$9b}
\item \texttt{huggingfacem4$\_$idefics$\_$9b$\_$instruct}
\item \texttt{llava$\_$1.6$\_$mistral$\_$7b}
\item \texttt{llava$\_$1.6$\_$vicuna$\_$13b}
\item \texttt{llava$\_$1.6$\_$vicuna$\_$7b}
\item \texttt{microsoft$\_$llava$\_$1.5$\_$13b$\_$hf}
\item \texttt{microsoft$\_$llava$\_$1.5$\_$7b$\_$hf}
\item \texttt{mistralai$\_$bakllava$\_$v1$\_$hf}
\item \texttt{openai$\_$gpt$\_$4$\_$1106$\_$vision$\_$preview}
\item \texttt{openai$\_$gpt$\_$4$\_$vision$\_$preview}
\item \texttt{openai$\_$gpt$\_$4o$\_$2024$\_$05$\_$13}
\item \texttt{openflamingo$\_$openflamingo$\_$9b$\_$vitl$\_$mpt7b}
\item \texttt{qwen$\_$qwen$\_$vl}
\item \texttt{qwen$\_$qwen$\_$vl$\_$chat}
\item \texttt{writer$\_$palmyra$\_$vision$\_$003}
\end{enumerate}
}

\newpage

\clearpage

\section{Sample-level Rankings}
\label{appx:rankings}

\subsection{Further Details}
In our ONEBench formulation, $s_j \in \mathcal{S}$ represents an ordinal ranking over the models $\mathcal{M}_j$ for sample $(x_j,y_j)$ 
represented by a permutation $\sigma_j$ such that $f_{\sigma_j(1)} \succeq \dots \succeq f_{\sigma_j(m_j)}$ where $m_j=|\mathcal{M}_j|$ is the number of models compared in the $j$-th sample-level ranking. In addition, for each $k$ we distinguish the case $f_{\sigma(k-1)} \succ f_{\sigma(k)}$ if $f_{\sigma(k-1)}$ performs better than $f_{\sigma(k)}$ and $f_{\sigma(k-1)} \sim f_{\sigma(k)}$  in case of indistinguishable performance. Thus, each sample-level ranking $s_j \in \mathcal{S}$ can be uniquely determined by a mapping $\sigma_j: \{1, \dots, m_j\} \to \{1, \dots, m\}$ with $\sigma_j(k)$ providing the index of the model in $\mathcal{M}$ that is on the $k$-th place in the ordering for the $j$-th sample-level ranking  and $\pi_j \in \{\succ, \sim \}^{m_j-1}$ defining the corresponding binary sequence of pairwise performance relations.\vspace{0.15cm}

\subsection{Ordinal Rankings and Information Loss} Using ordinal measurements leads to information loss, which can impede downstream aggregation algorithms due to the data processing inequality (\citealt{thomas2006elements}, Section 2.8). This principle asserts that any estimation made from processed data cannot outperform estimation based on the original, unprocessed data. However, cardinal measurements frequently suffer from calibration issues, even within a single metric \citep{shah2014better}. Consequently, in practice, ordinal measurements can paradoxically outperform cardinal ones despite the inherent information loss.\vspace{0.15cm}
\newpage
\section{Synthetic data experiments}
\label{appx:synthetic}
To further verify the ability of our method to recover the ground truth ranking, we create a simulation of the data-generating process. Following~\citet{colombo2022best}, we generate synthetic scores using Gumbel-distributed random variables. Specifically, we simulate $N=100$ systems, each modeled by a Gumbel random variable $G_n$ centered at $\phi \cdot n$ with scale $\beta=1$, where $\phi \in [0,1]$ is a dispersion parameter. We then generate $n=1000$ measurements, where the score of each system is sampled from its corresponding Gumbel distribution. The measurements are in equal proportions numerical (raw scores), binary (whether the score is over a randomly selected threshold), and ordinal (whether the score of one system is higher than another). In Table~\ref{tab:noise_levels}, we present the Kendall rank correlation to the ground truth ranking for different values of $\phi$, which controls the noise in the ranking (a lower value is more noise).

\begin{table*}[htbp]
\centering

\begin{minipage}{0.49\textwidth}
\centering
\resizebox{!}{1.25cm}{%
\begin{tabular}{lccc}
\toprule
\textbf{Dispersion} $\boldsymbol{\phi}$ & \textbf{ELO} & \textbf{LMArena} & \textbf{Ours} \\
\midrule
0.01 & 0.28 ± 0.02 & 0.88 ± 0.00 & \textbf{0.92 ± 0.00} \\
0.02 & 0.39 ± 0.05 & 0.92 ± 0.01 & \textbf{0.96 ± 0.01} \\
0.05 & 0.74 ± 0.01 & 0.95 ± 0.00 & \textbf{0.98 ± 0.00} \\
0.10 & 0.85 ± 0.03 & 0.93 ± 0.01 & \textbf{0.99 ± 0.00} \\
\bottomrule
\end{tabular}}
\caption{Kendall rank correlation to the ground truth ranking at different dispersion levels of synthetic model measurements. Lower dispersion means more noise.}
\label{tab:noise_levels}
\end{minipage}
\hfill
\begin{minipage}{0.49\textwidth}
\centering
\resizebox{!}{1.25cm}{%
\begin{tabular}{lccc}
\toprule
\textbf{Missing data} & \textbf{ELO} & \textbf{LMArena} & \textbf{Ours} \\
\midrule
50\% & 0.87 ± 0.02 & 0.92 ± 0.00 & \textbf{0.99 ± 0.00} \\
90\% & 0.84 ± 0.01 & 0.90 ± 0.00 & \textbf{0.97 ± 0.00} \\
95\% & 0.84 ± 0.03 & 0.88 ± 0.01 & \textbf{0.95 ± 0.01} \\
99\% & 0.86 ± 0.01 & 0.83 ± 0.01 & \textbf{0.89 ± 0.01} \\
\bottomrule
\end{tabular}}
\caption{Kendall rank correlation to the ground truth ranking with different percentages of synthetic model measurements missing ($\phi=0.1$).}
\label{tab:missing_data}
\end{minipage}
\end{table*}

\newpage

\section{Separability}
\label{appx:separability}
To study separability, we follow the definition in~\citet{noothigattu2020axioms} and randomly split the data in each benchmark into two equal-sized subsets 5 times. Then, we use all three aggregation methods to produce sub-rankings, as well as a combined ranking. We identify model pairs that are consistently ranked in both sub-rankings and compute the percentage of cases where that ordering is maintained in the combined ranking. The results, showing the mean and standard deviation of these percentages, are presented in Table~\ref{tab:separability_results}.

\begin{table}[h!]
\centering
\resizebox{0.8\linewidth}{!}{  
\begin{tabular}{lccc}
\toprule
\textbf{Benchmark} & \textbf{ELO (mean ± std)} & \textbf{LMArena (mean ± std)} & \textbf{Plackett-Luce (mean ± std)} \\
\midrule
HELM        & 91.84 ± 0.25  & 100.00 ± 0.00 & 100.00 ± 0.00 \\
Leaderboard & 83.11 ± 0.73  & 100.00 ± 0.00 & 100.00 ± 0.00 \\
VHELM       & 95.81 ± 0.77  & 100.00 ± 0.00 & 100.00 ± 0.00 \\
LMMs-Eval   & 91.85 ± 0.88  & 100.00 ± 0.00 & 100.00 ± 0.00 \\
Synthetic   & 97.37 ± 0.40  & 100.00 ± 0.00 & 100.00 ± 0.00 \\
\bottomrule
\end{tabular}
}
\caption{\textbf{Separability analysis:} Percentage of consistent pairwise rankings from sub-rankings preserved in the combined ranking.}
\label{tab:separability_results}
\end{table}

\newpage

\section{Capability Testing Across Arbitrary Queries}
\label{appx:capability_probing}
\subsection{Queries: List and Additional Results}
\label{appx:capability_quant}
\begin{table*}[h]
\centering
\renewcommand{\arraystretch}{0.55}
\resizebox{0.8\textwidth}{!}{%
\begin{tabular}{l|c|c}
\toprule
\textbf{Concept} & ONEBench-LLM AP & ONEBench-LMM AP \\
\hline
\multicolumn{3}{c}{Common Queries}\\ \hline
apple ipad & 0.7435 & 0.1985 \\
architecture & 0.7683 & 0.8981 \\
beach & 0.7152 & 0.5698 \\
biochemistry & 0.9778 & 0.7303 \\
boat & 0.7728 & 0.8829 \\
botany & 0.9876 & 0.7556 \\
bus & 0.9035 & 0.9739 \\
car & 0.9140 & 0.8477 \\
cell(biology) & 0.9937 & 0.5075 \\
china tourism & 0.6392 & 1.0000 \\
cigarette advertisment & 0.7249 & 0.6590 \\
coffee maker & 0.8426 & 0.4057 \\
components of a bridge & 0.9222 & 0.5865 \\
decomposition of benzene(organic chemistry) & 0.6745 & 0.7623 \\
epidemiology & 0.9316 & 0.7991\\
kirchoff's law(electrical engineering)  & 0.6572 & 0.4824 \\
food chain & 0.5405 & 1.0000 \\
game of football & 0.8221 & 1.0000 \\
german shepherd (dog) & 0.9359 & 0.3078 \\
gothic style (architecture) & 0.7829 & 1.0000 \\
law & 0.8566 & 0.4138\\
literary classics & 0.9869 & 1.0000 \\
macroeconomics & 1.0000 & 0.9570 \\
makeup & 1.0000 & 0.2247 \\
microwave oven & 0.7979 & 1.0000 \\
neuroscience components & 0.9844 & 0.2854 \\
pasta & 0.5678 & 0.2142 \\
perfume & 0.5996 & 0.6355 \\
photosynthesis & 0.9848 & 0.3665 \\
plants & 1.0000 & 0.6488 \\
political diplomacy & 0.9529 & 0.9561 \\
python code & 0.8850 & 0.9444 \\
renaissance painting & 0.9270 & 0.9799 \\
shareholder report & 1.0000 & 0.8317 \\
sheet music & 0.8322 & 0.9750 \\
solar cell battery & 0.8853 & 0.8082 \\
thermodynamics & 0.9567 & 0.8852 \\
united states of america & 0.8096 & 0.8642 \\
vaccines & 0.8572 & 0.3411\\
volcanic eruption & 0.7905 & 0.9229 \\
\hline
\multicolumn{3}{c}{Queries testing Visual Capabilities}\\\hline
bike leaning against a wall & - & 0.8271 \\
child playing baseball & - & 0.9638\\
coriolis effect & - & 0.7063 \\
dijkstra's shortest path algorithm & - & 0.9135\\
empty bridge overlooking the sea & - & 0.5934 \\
judo wrestling & - & 0.6092 \\
man in a suit & - & 0.5611\\
musical concert & - & 0.9879 \\
sine wave & - & 0.4232 \\
woman holding an umbrella & - & 0.8821\\
\bottomrule
\end{tabular}
}
\caption{Aggregate Average Precision(AP) for ONEBench-LLM and ONEBench-LMM concepts.}
\label{tab:ap_comparison}
\end{table*}
\begin{figure*}[t!]
    \centering
    \includegraphics[width=\linewidth]{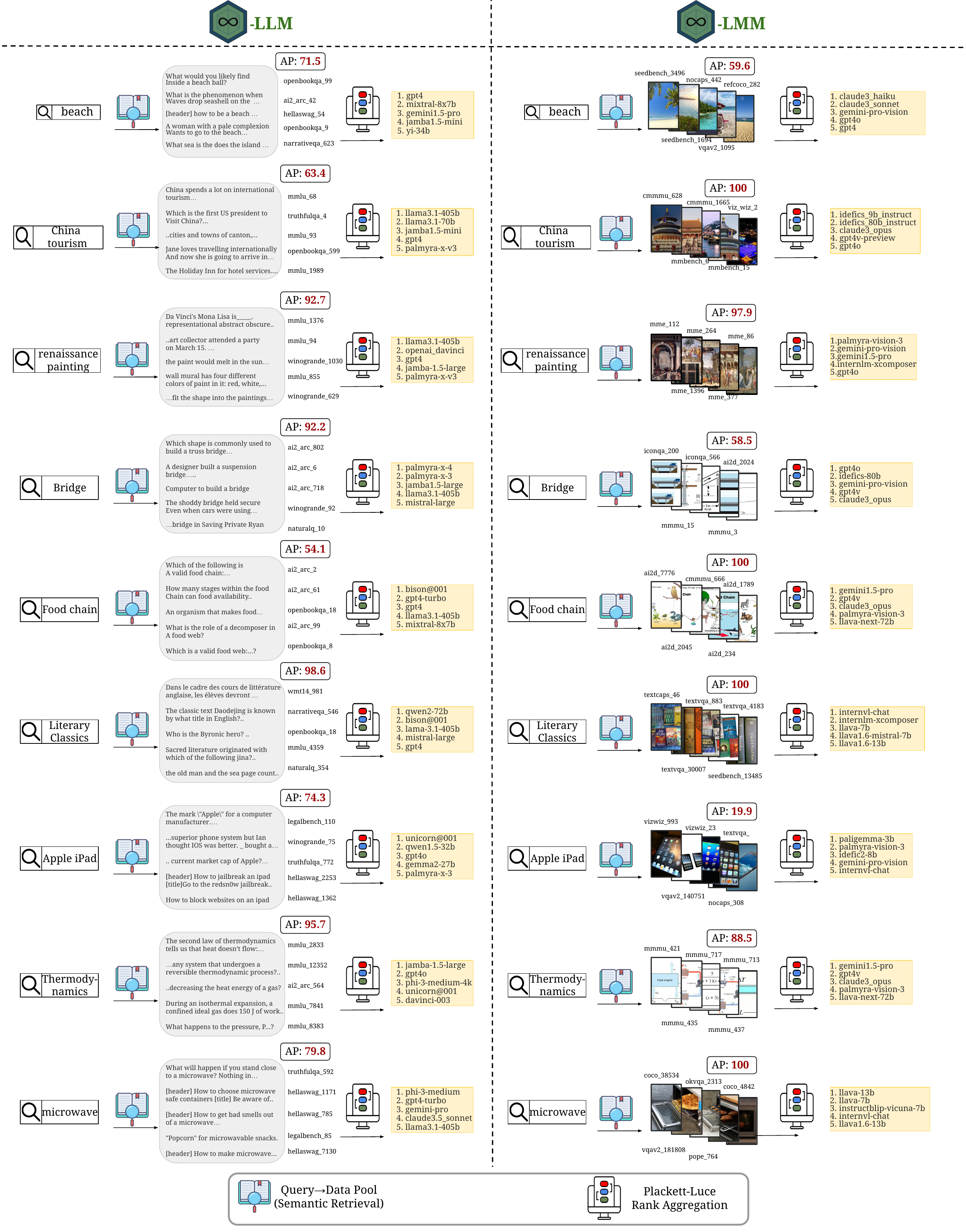}
    \caption{Additional qualitative analysis for ONEBench's capability probing for selected queries.}
    \label{fig:capability_probing_appendix}
\end{figure*}

\newpage
\clearpage

\subsection{Variability in Concept-specific Rankings}
We extend the analysis conducted in Section \ref{subsec:probing} demonstrating that the performance gap is non-trivial while comparing model rankings for specific capabilities to randomly sampling datapoints and aggregating model performance. Showcasing this reiterates the practical value of ONEBench - such that practioners who are interested in specific capabilities may extract the best LLM/LMM for their needs. 

We already studied how the number of samples determines model ranking variation over the entire concept pool. Now, we check whether the top-k model rankings vary between the global ranking over the whole data pool and for specific capabilities.

Regarding the top-k models, we do a specific study into ONEBench-LMM, as we test on more concepts than ONEBench-LMM(50 vs 40). The overall top-1 model using the Plackett-Luce algorithm, evidenced as the state-of-the-art rank aggregation method, combining both LMMs-Eval and VHELM is \texttt{openai\_gpt\_4o\_2024\_05\_13}. Hence, we conduct our analysis on the top-5 and top-10 model rankings.

For all the concepts collected for ONEBench-LMM (Table \ref{tab:ap_comparison}), we provide aggregated results of correlations between the overall ranking obtained by the Plackett-Luce algorithm and specific benchmark rankings. We indicate the overall rank differences for each concept with respect to the overall ranking and the average Kendall-$\tau$ correlation of model rankings for all the capability-specific benchmarks. We also conduct a Wilcoxon signed-rank test as a hypothesis test to check the systematic difference between paired observations as shown in Table \ref{tab:hypothesis_test}.

\begin{table}[h]
\centering
\begin{tabular}{lcc}
\toprule
\textbf{Metric} & \textbf{Top-5} & \textbf{Top-10} \\
\midrule
Mean absolute rank difference & 7.900 & 7.920 \\
Mean Kendall's $\tau$ correlation & 0.220 & 0.124 \\
Wilcoxon test statistic & 53.000 & 623.000 \\
Wilcoxon $p$-value & $9.66 \times 10^{-7}$ & $1.63 \times 10^{-8}$ \\
Conclusion & Significant difference & Significant difference \\
\bottomrule
\end{tabular}
\caption{Comparison of Overall vs. Capability-Specific Rankings}
\label{tab:hypothesis_test}
\end{table}

We infer from the results that there is a 
 significant difference between overall and specific rankings. Hence, across experiments, we observe variations across the capabilities we test and conclude that we cannot have a one-model-fits-all system for different concepts and domains.

 \newpage
\section{Extended Related Works}
\label{appx:rw}
\textbf{Multi-task Benchmarks as Broad Capability Evaluators.} Multi-task leaderboards have been the standard for benchmarking foundation models. Examples include GLUE~\citep{wang2018glue}, decaNLP \citep{mccann2018natural}, SuperGLUE~\citep{wang2019superglue},  BigBench~\citep{srivastava2022beyond}, Dynabench~\citep{kiela2021dynabench}, Open LLM Leaderboard~\citep{open-llm-leaderboard}, CLIP-Benchmark~\citep{LAION_CLIP_benchmark}, ELEVATOR~\citep{li2022elevater}, StableEval~\citep{udandarao2024active} and DataComp-38~\citep{gadre2023datacomp}, as well as massive multitask benchmarks like XTREME \citep{siddhant2020xtreme} and ExT5 \citep{aribandi2021ext5}. However, concerns have arisen regarding the limitations of multi-task benchmarks \citep{bowman2021will}. Issues include saturation and subsequent discarding of samples \citep{liao2021we, beyer2020we, ott2022mapping, ethayarajh2020utility, xia2024top}, susceptibility to dataset selection \citep{dehghani2021benchmark}, obscuring progress by evaluation metrics \citep{schaeffer2024emergent, colombo2022infolm}, training on test tasks \citep{udandarao2024no, dominguez2024training, nezhurina2024alice, mirzadeh2024gsm, srivastava2024functional, wang2024benchmark}, and data contamination \citep{elangovan2021memorization, magar2022data, deng2023investigating, golchin2023data, sainz2024data}. ONEBench tackles these challenges by enabling extensive reuse of samples for broader model comparisons, avoiding task selection bias through democratized sourcing of samples, and using ordinal rankings to avoid evaluation minutia. Sample-level evaluation with sparse inputs also allows selective removal of contaminated data for fairer comparisons. Moreover, by supporting over-ended, evolving evaluation, it makes it harder to train on all test tasks, as opposed to fixed leaderboards that are easier to game.\vspace{0.15cm}

\noindent \textbf{On Aggregation across Benchmarks.} The dominant approach to benchmarking has traditionally been multi-task benchmarks, where the most common aggregation strategy is the arithmetic mean of scores across individual tasks. However, this approach assumes that the scoring metrics are homogeneous and scaled correctly, and treat tasks of different complexities equally \citep{mishra2021robust, pikuliak2023average}. In consequence, simple normalization preprocessing influences the rankings \citep{colombo2022best}, and makes them nearly entirely dependent on outlier tasks \citep{agarwal2021deep}. Simply changing the aggregation method from arithmetic to geometric or harmonic mean can change the ranking \citep{shavrina2021not}. Similarly, including irrelevant alternative models can change statistical significance or even change the ranking entirely \citep{benavoli2016should, zhang2024inherent}. Mean-aggregation also has significant failure modes in handling missing scores in benchmarks \citep{himmi2023towards}. The benchmarking paradigm is hence shifting towards adopting evaluation principles from other fields, such as non-parametric statistics and social choice theory \citep{brandt2016introduction, rofin2022vote}. We use ordinal rankings instead of scores, similar to LMArena. However, unlike Arena, we use the pairwise variant of the Plackett-Luce model, which has been shown to have advantages both theoretically and empirically \citep{peyrard2021better}. We benefit from some of its theoretical properties like identifiability, sample-efficient convergence, provable robustness to irrelevant alternatives, non-dominance of outliers and empirical robustness across a wide range of real-world factors which affect ranking. Moreover, we do not aggregate over benchmarks in the first place---our primary proposal is to avoid monolithic benchmarks and consider aggregation on a sample level, needing to tackle incomplete and heterogeneous measurements. We note that several other social-choice theory-based models such as score-based models \citep{shevchenko2024variability} based on the Condorcet-winner criterion \citep{young1988condorcet} have been proposed, yet they were primarily applied for aggregation on multi-task benchmarks, whereas a crucial component of our proposal is to break down the benchmark boundaries and aggregate heterogeneous samples.\vspace{0.15cm}

\textbf{Dynamic Evaluation and Active Testing.} Some previous works like~\cite{ji2021active,kossen2021active,kossen2022active,saranathan2024dele,huang2024active, zhu2023dyval} tackle the `active testing' problem, where the goal is to identify small ``high-quality'' test data-subsets, from a large pool of uncurated evaluation data.
These works typically assume that the cost of unlabeled test data acquisition is low whereas the cost of acquiring per-instance labels is high. However, as pointed out by~\citet{prabhu2024lifelong}, these assumptions are unrealistic for foundation models, as both the acquisition of test data and label annotations can be tedious in general. Hence, in our work, we tackle a broader problem: given a large testing data pool, how can we curate and query to produce a consistent and targeted set of model rankings?
\vspace{0.15cm}

\textbf{Efficient Evaluation.} As evaluation suites have grown, associated inference costs have also increased. Recent research has focused on creating compressed subsets of traditional benchmarks to address this issue \citep{varshney2022ildae, zhao2024flasheval,perlitz2023efficient,kipnis2024texttt,pacchiardi2024100}. Popular approaches include subsampling benchmarks to preserve correlations with an external source like LMArena~\citep{ni2024mixeval}, sample clustering to gauge sample difficulty and then sub-sampling~\citep{vivek2023anchor}, item-response-theory based methods for informatively sampling a subset of samples for evaluation~\citep{polo2024tinybenchmarks}, or designing evolving sample-level benchmarks~\citep{prabhu2024lifelong}. While the work of~\citet{prabhu2024lifelong} is similar to us in principle, it requires binary metrics as input and does not handle incomplete input matrices, which is necessary for aggregation over multiple time steps. We precisely address these limitations by showing efficient evaluation while accommodating incomplete data and extending it to ordinal ranks.\vspace{0.15cm}

\textbf{Democratizing Evaluation.} Standard image classification and retrieval benchmarks are collected from platforms like Flickr, which are predominantly Western-centric~\citep{ananthram2024see,shankar2017no}. This has raised the important question: ``Progress for whom?'', with many seminal works showcasing large disparities in model performance on concepts~\citep{hemmat2024improving}, tasks~\citep{hall2024towards,hall2023dig,hall2023towards}, and even input samples~\citep{pouget2024no,sureddy2024decomposed,gustafson2024exploring} from the Global South. In response, works have developed benchmarks tailored to diverse cultures and demographics to include their voice in measuring progress~\citep{pistilli2024civics, pouget2024no, nguyen2024multilingual, luccioni2023bugs}. Further works have tried to create personalized, task-specific benchmarks for flexibly evaluating models based on user-preferences~\citep{butt2024benchagents,saxon2024benchmarks, yuan2024s, li2024autobencher}---~\citet{zhang2024task} created Task-Me-Anything that enables users to input specific queries that then get processed to provide model rankings or responses to the query. However, their system is entirely procedurally generated, thereby not reflecting the real-world use-cases that models are typically subjected to in practice. Further, they are restricted to the fixed set of instances in their task generator pool. We take a different approach by creating flexible benchmarks where individuals, and contributing entities, can add their own samples and preferences collected from both real-world benchmarks and live model arenas like LM-Arena, thereby providing users with a realistic overview of model rankings on practical scenarios.
Further, during capability testing, users can select similar preferences, making ONEBench more inclusive than traditional test sets.

\clearpage
\newpage
\section{Open Problems and Future Directions}
In this section, we highlight some promising directions
for improvement below:

\begin{enumerate}[leftmargin=*]
    \item \textcolor{kleinblue2}{\underline{Testing Limits and Scaling Up ONEBench:}} currently, our prototype comprises less than 100K samples in ONEBench-LLM and under 1M in ONEBench-LMM. These pools can be greatly expanded and diversified by expanding to incorporating \textit{all existing} LLM and LMM benchmarks. Our retrieval mechanisms are designed to scale efficiently as the test pool grows in size and diversity.
    \item \textcolor{kleinblue2}{\underline{Exploring Other Aggregation Algorithms:}} while we use the Plackett-Luce model for aggregating diverse measurements, there exist other algorithms from computational social choice theory with different trade-offs. A comprehensive evaluation of these alternatives could offer new insight for aggregating model performance.
    \item \textcolor{kleinblue2}{\underline{Structured Querying and Enhanced Retrieval:}} One can improve retrieval by better querying mechanisms using models like ColBERT~\citep{khattab2020colbert} and ColPALI~\citep{faysse2024colpali}, further optimized using DSPy~\citep{khattab2023dspy}. A particularly interesting direction is allowing compositional queries, where users combine multiple queries to test behaviour in foundation models, similar to works like ConceptMix~\citep{wu2024conceptmix} and SkillMix~\citep{yu2023skill}. 

    \item \textcolor{kleinblue2}{\underline{On the Limits of Capability Probing:}} While we currently allow broad, open-ended inputs to probe capabilities, some are easier to assess than others~\citep{madvil2023read,li2024crowdsourced}. As foundation models become more generalizable, a thorough analysis identifying which capabilities can be \textit{easily, reliably evaluated}, which are \textit{possible to evaluate but challenging}, and which are in principle \textit{impossible to evaluate} is needed---this will help improve benchmarking effectiveness.
\end{enumerate}
\newpage

\end{document}